\documentclass[journal]{IEEEtran}

\usepackage{cite}

\ifCLASSINFOpdf
  \usepackage[pdftex]{graphicx}
\else
  \usepackage[dvips]{graphicx}
\fi

\usepackage[cmex10]{amsmath}
\usepackage{subfigure}
\usepackage{flushend}
\usepackage{amssymb}
\usepackage{multirow}
\usepackage{makecell}
\usepackage{marvosym}
\usepackage{url}
\usepackage{algorithm}
\usepackage{algorithmic}

\graphicspath{{Figure/}}

\begin{document}

\title{Joint Anchor-Feature Refinement for Real-Time Accurate Object Detection in Images and Videos}

\author{Xingyu Chen, Junzhi Yu, \emph{Senior Member, IEEE}, Shihan Kong, Zhengxing Wu, and Li Wen, \emph{Member, IEEE} 

\thanks{This work was supported in part by the National Key Research and Development Program of China under Grant 2019YFB1310300 and by the National Natural Science Foundation of China under Grant 61633004. \emph{(Corresponding author: Junzhi Yu.)}}
\thanks{X. Chen, S. Kong, and Z. Wu  are with the State Key Laboratory of Management and Control for Complex Systems, Institute of Automation, Chinese Academy of Sciences, Beijing 100190, China, and also with the School of Artificial Intelligence, University of Chinese Academy of Sciences, Beijing 100049, China (e-mail: chenxingyu2015@ia.ac.cn; kongshihan2016@ia.ac.cn; zhengxing.wu@ia.ac.cn).}
\thanks{J.~Yu is with the State Key Laboratory of Management and Control for Complex Systems, Institute of Automation, Chinese Academy of Sciences, Beijing 100190, China, and also with the State Key Laboratory for Turbulence and Complex Systems, Department of Mechanics and Engineering Science, BIC-ESAT, College of Engineering, Peking University, Beijing 100871, China (e-mail: junzhi.yu@ia.ac.cn).}
\thanks{L. Wen is with the School of Mechanical Engineering and Automation, Beihang University, Beijing 100191, China (e-mail: liwen@buaa.edu.cn).}
}

\maketitle

\begin{abstract}
Object detection has been vigorously investigated for years but fast accurate detection for real-world scenes remains a very challenging problem. Overcoming drawbacks of single-stage detectors, we take aim at precisely detecting objects for static and temporal scenes in real time. Firstly, as a dual refinement mechanism, a novel anchor-offset detection is designed, which includes an anchor refinement, a feature location refinement, and a deformable detection head. This new detection mode is able to simultaneously perform two-step regression and capture accurate object features. Based on the anchor-offset detection, a dual refinement network (DRNet) is developed for high-performance static detection, where a multi-deformable head is further designed to leverage contextual information for describing objects. As for temporal detection in videos, temporal refinement networks (TRNet) and temporal dual refinement networks (TDRNet) are developed by propagating the refinement information across time. We also propose a soft refinement strategy to temporally match object motion with the previous refinement. Our proposed methods are evaluated on PASCAL VOC, COCO, and ImageNet VID datasets. Extensive comparisons on static and temporal detection verify the superiority of DRNet, TRNet, and TDRNet. Consequently, our developed approaches run in a fairly fast speed, and in the meantime achieve a significantly enhanced detection accuracy, i.e., 84.4\% mAP on VOC 2007, 83.6\% mAP on VOC 2012, 69.4\% mAP on VID 2017, and 42.4\% AP on COCO. Ultimately, producing encouraging results, our methods are applied to online underwater object detection and grasping with an autonomous system. Codes are publicly available at \url{https://github.com/SeanChenxy/TDRN}.
\end{abstract}

\section{Introduction}
\label{sec:Int}
Object detection is one of the fundamental and challenging areas of research in computer vision. With rapid advances in deep learning, convolutional neural networks (CNN) have demonstrated the state-of-the-art performance in this task. Zhao \emph{et al.} presented an overview of modern object detection approaches \cite{bib:Review}. From this review, we can see that two-stage detectors represented by RCNN family \cite{bib:FasterRCNN} and RFCN \cite{bib:RFCN} usually attain an accurate yet slightly slow performance. On the contrary, by detecting objects in a one-step fashion, single-stage detectors \cite{bib:YOLO,bib:SSD} are able to run in real time with reasonably modest accuracy. Therefore, fast accurate detection remains a challenging problem for real-world applications.

\begin{figure}[!t]
 \centering
\includegraphics[width=8cm]{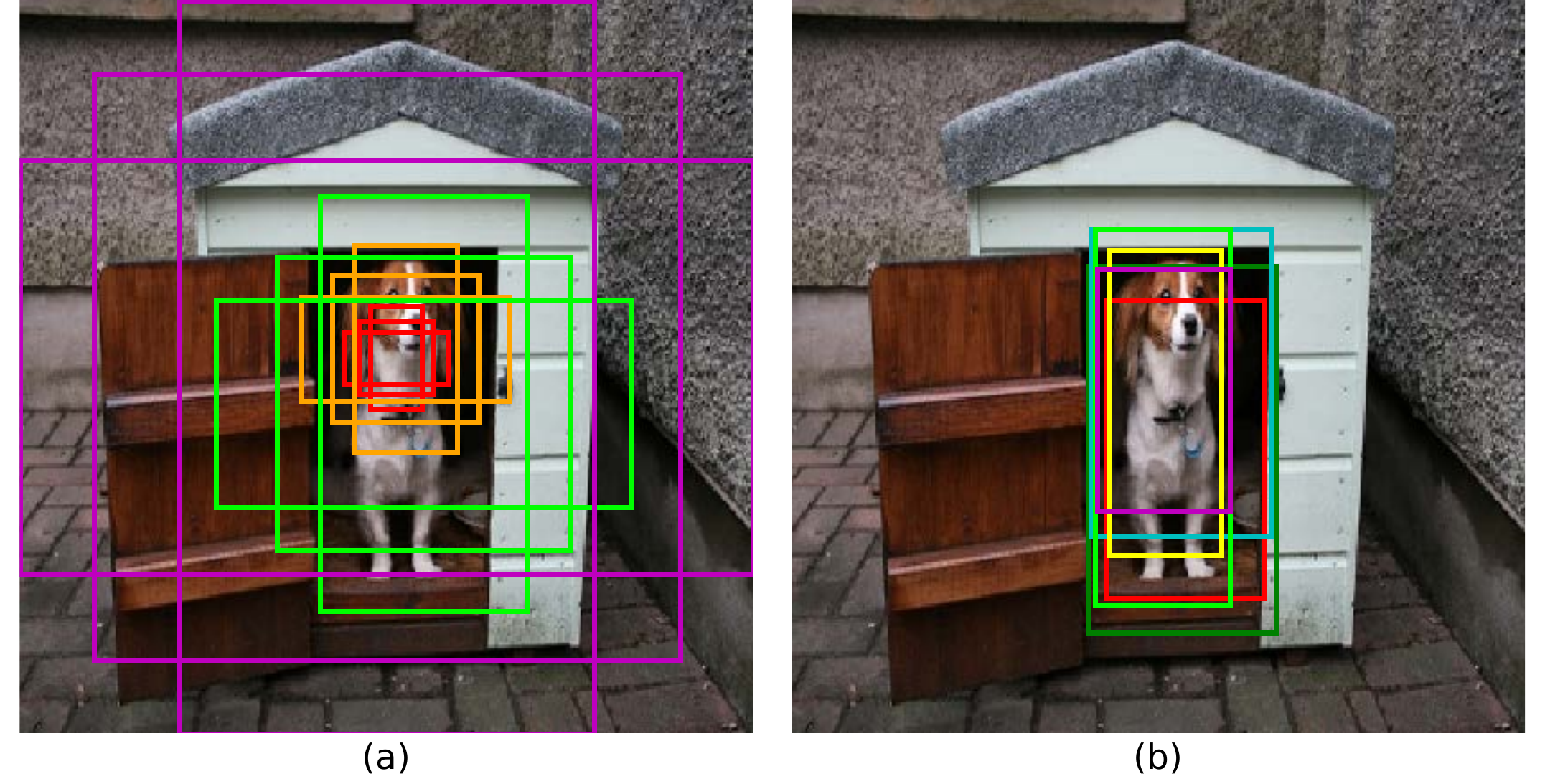} 
\caption{Comparison of single-stage anchors and RPN outputs. For better visualization, only several key boxes are demonstrated. (a) Multi-scale SSD anchors. (b) RPN outputs in Faster RCNN.}
\label{fig:intro}
\end{figure}

It is instructive that the two-stage method induces high accuracy, while the single-stage detector has a desirable inference speed. This inspires us to investigate the reasons. In our opinion, the high accuracy of two-stage approaches comes with two advantages: i) two-step regression and ii) relatively accurate features for detection. In detail, two-stage detectors firstly regress pre-defined anchors with the aid of region proposal \cite{bib:FasterRCNN}, and this operation significantly eases the difficulty of final localization. Besides, an RoI-wise subnetwork \cite{bib:FasterRCNN} is appended to the region proposal part, so features in region of interest can be leveraged for final detection. By contrast, there are two drawbacks in the single-stage paradigm: i) Detection head directly regresses coordinates from pre-defined anchors, but most anchors are far from matching object regions. ii) Classification information comes from probably inaccurate locations, where features could not be precise enough to describe objects. Referring to Fig.~\ref{fig:intro}(a), it is relatively difficult to regress pre-defined anchors to precisely surround the object (e.g., the dog in Fig.~\ref{fig:intro}). Moreover, as feature sampling locations follow pre-defined anchor regions, detection features for small-scale anchors cannot cover the entire object region, while that for large-scale anchors weaken the object because of background. On the contrary, owing to region proposal, the two-stage methods detect the dog using a better initialization (see Fig.~\ref{fig:intro}(b)). Thus, the strengths of two-stage methods exactly reflect the single-stage drawbacks that lead to relatively lower detection accuracy. Although Zhang \emph{et al.} developed RefineDet \cite{bib:RefineDet} to introduce two-step regression to the single-stage detector, it still failed to capture accurate detection features. That is, pre-defined feature sampling locations are not precise enough for describing refined anchor regions. (Note that detailed comparison between RefineDet and our approach will be presented in Section~\ref{sec:AOD}.) Thus, there is an imperative need of further overcoming these single-stage limitations for real-time accurate object detection.

In addition, most researches have largely focused on detecting object statically, ignoring temporal coherence in real-world applications. Detection in real-world scenes was introduced by ImageNet video detection (VID) dataset \cite{bib:ImageNet}. To the best of our knowledge, main ideas of temporal detection include i) post-processing \cite{bib:SeqNms}, ii) tracking-based location \cite{bib:TCNN,bib:DT}, iii) feature aggregation with motion information \cite{bib:TCNN,bib:FGFA,bib:HPVD,bib:HPVD_Mobiles,bib:STSN}, iv) RNN-based feature propagation \cite{bib:TSSD,bib:STMN,bib:LSTM_SSD,bib:HPVD_Mobiles}, and v) batch-frame processing (i.e., tubelets proposal) \cite{bib:TPN}. All these ideas are attractive in that they are able to leverage temporal information for detection, but they also have respective limitations. In brief, methods i)--iv) borrow other tools (e.g., tracker, optical flow, LSTM, etc.) for temporal analysis. Methods iii) and iv) focus on constructing superior temporal features. Nevertheless, they detect objects following the static mode. Method v) works in a non-causal offline mode that prohibits this approach from real-world tasks. Furthermore, most recent works pay excessive attention to accuracy so that high computational costs could affect time efficiency. Thus, a novel temporal detection mode should be developed for videos.

Overcoming aforementioned single-stage drawbacks, a dual refinement mechanism is proposed in this paper for static and temporal visual detection, namely anchor-offset detection. This joint anchor-feature refinement includes an anchor refinement, a feature location refinement, and a deformable detection head. The anchor refinement is developed for two-step regression, while the feature location refinement is proposed to capture accurate single-stage features. Besides, a deformable detection head is designed to leverage this dual refinement information. Based on the anchor-offset detection, we propose three approaches for object detection in images and videos. Firstly, a dual refinement network (DRNet) is proposed. DRNet is designed for static detection, where a multi-deformable head is developed for diversifying detection receptive fields for more contextual information. Secondly, temporal refinement networks (TRNet) are designed, which perform anchor refinement across time for video detection. Thirdly, temporal dual refinement networks (TDRNet) are developed that extend the anchor-offset detection towards temporal tasks. Additionally, for temporal detection task, we propose a soft refinement strategy to match object motion with previous refinement information. Our proposed DRNet, TRNet, and TDRNet are validated on PASCAL VOC \cite{bib:VOC}, COCO \cite{bib:COCO}, and ImageNet VID \cite{bib:ImageNet} datasets. As a result, our methods achieve a real-time inference speed and considerably improved detection accuracy. Furthermore, these approaches have been applied to object-driven navigation and grasping in an unstructured undersea environment. Contributions are summarized as follows:
\begin{itemize}
  \item Starting with drawbacks of single-stage detectors, an anchor-offset detection is proposed to perform two-step regression and capture accurate object features. The anchor-offset detection includes an anchor refinement, a feature-offset refinement, and a deformable detection head. Academically, without region-level processing, this joint anchor-feature refinement achieves single-stage region proposal. Thus, the anchor-offset detection bridges single-stage and two-stage detection so that it is able to induce a new detection mode.
  \item A DRNet based on the anchor-offset detection and a multi-deformable head is developed to elevate static detection accuracy while maintaining real-time inference speed for image detection task.
  \item As a new temporal detection mode for video detection task, a TRNet and a TDRNet are proposed based on the anchor-offset detection without the aid of any other temporal modules. They are characterized by a better accuracy vs. speed trade-off and have a concise training process without the requirement of sequential data. In addition, a soft refinement strategy is designed to enhance the effectiveness of refinement information across time.
  \item The single-stage DRNet maintains fast speed while acquiring significant improvements in accuracy, i.e., $84.4\%$ mean average precision (mAP) on VOC 2007 test set, $83.6\%$ mAP on VOC 2012 test set, and $42.4\%$ AP on COCO test-dev. Based on VID 2017 validation set, DRNet sees $69.4\%$ mAP; TRNet achieves $66.5\%$ mAP; and TDRNet obtains $67.3\%$ mAP.
\end{itemize}

The remainder of this paper is organized as follows. Section~\ref{sec:RW} presents the related works. Including anchor-offset detection and multi-deformable head, DRNet is elaborated in Section~\ref{sec:DRN}. Section~\ref{sec:TRN} presents TRNet and TDRNet in detail, and Section~\ref{sec:EXP} provides the experimental results and discussion. Conclusions are summarized in Section~\ref{sec:CON}.

\section{Related Work}
\label{sec:RW}

\begin{figure*}[!t]
 \centering
\includegraphics[width=18cm]{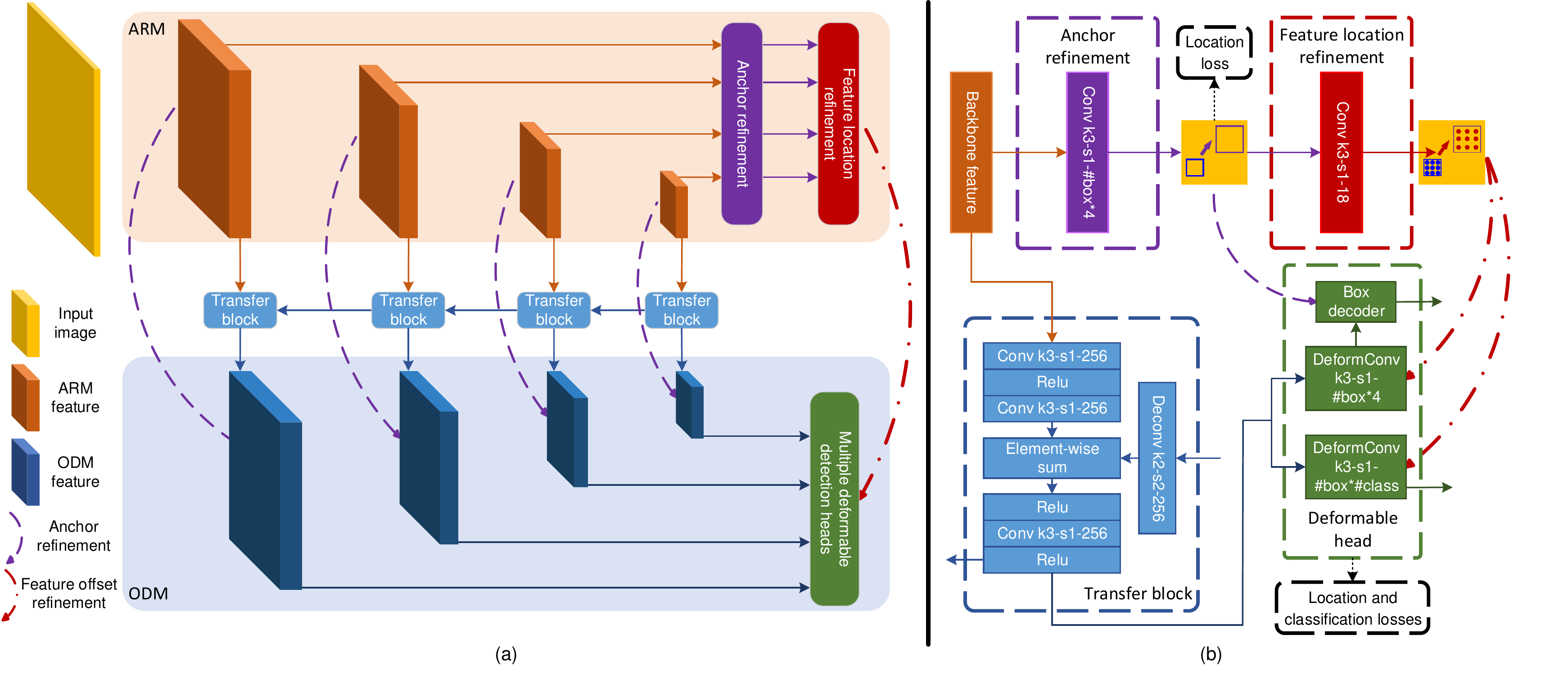}
\caption{The schematic layout of the proposed DRNet. Refined anchors are produced by coarse regression with ARM features, and they are first employed to predict feature offsets, namely, feature location refinement. The detection head utilizes ODM feature maps, refined anchors, and refined feature sampling locations to detect objects, i.e., anchor-offset detection. A multi-deformable head is designed for rich contextual information. (a) Overall framework. Each anchor refinement induces a feature location refinement. (b) Design details. Convolution is detailed with kernel size, stride, and output channel size. Only one detection path with $3\times3$ convolutional kernel is shown.}
\label{fig:DRN}
\end{figure*}

\subsection{CNN-Based Static Object Detection}
Deep learning methods have recently dominated the field of object detection \cite{bib:Review}. Two-stage detectors \cite{bib:FasterRCNN,bib:RFCN,bib:CoupleNet,bib:ACoupleNet} usually detect objects by region proposal, location, and classification. For example, inspired by Faster RCNN \cite{bib:FasterRCNN} and RFCN \cite{bib:RFCN}, CoupleNet \cite{bib:CoupleNet} leveraged both region-level and part-level features to express a variety of challenging object situations, which achieved considerable detection accuracy but it only ran at $8.2$ FPS. As groundbreaking works, YOLO \cite{bib:YOLO} and SSD \cite{bib:SSD} localized and classified objects using a single-shot network for real-time detection. Recently, many revised single-stage versions have emerged \cite{bib:DSSD,bib:RefineDet,bib:RetinaNet,bib:RFB,bib:YOLOv2,bib:DSOD,bib:YOLOv3}. Typically, in favor of small object detection, Lin~\emph{et al.} developed a RetinaNet to formulate the single-shot network as an FPN \cite{bib:FPN} fashion for propagating information in a top-down manner to enlarge shallow layers' receptive field \cite{bib:RetinaNet}. Redmon and Farhadi proposed YOLOv3 with DarkNet53 and multi-scale anchor for fast accurate detection \cite{bib:YOLOv3}. Zhang~\emph{et~al.} designed a RefineDet to introduce two-step regression to single-stage pipeline \cite{bib:RefineDet}. RefineDet adjusted pre-defined anchors for more precise localization. However, its detection features were still fixed on pre-defined positions, failing to precisely describe refined anchor regions. In short, although single-stage methods has a superiority in speed, two-stage methods still dominate detection accuracy on generic benchmarks \cite{bib:VOC,bib:COCO,bib:ImageNet}. Hence, we are motivated to analyze single-stage drawbacks from two-stage merits (analyzed in Section~\ref{sec:Int}), and construct DRNet with both competitive accuracy and fast speed.

\subsection{Temporal Object Detection}
To detect objects in temporal vision, some post-processing methods have been first investigated to merge multi-frame results, then tracker-based detection, motion-guided feature aggregation, RNN-based feature integration, and tubelet proposal are studied by the research community. Han \emph{et al.} proposed an SeqNMS to discarded temporally interrupted bounding boxes in the non-maximum suppression (NMS) phrase \cite{bib:SeqNms}; Feichtenhofer \emph{et al.} combined RFCN and a correlation-filter-based tracker to boost recall rate \cite{bib:DT}. Based on motion estimation with optical flow, Zhu \emph{et al.} devised a temporally-adaptive key frame scheduling to effective feature aggregation \cite{bib:HPVD}; Chen \emph{et~al.} and Liu \emph{et al.} took advantage of Long Short-Term Memory to propagate CNN features across time \cite{bib:TSSD,bib:LSTM_SSD}. However, the temporal analysis capacity in the above-mentioned methods is obtained from other temporal tools. Although some methods focused on how to construct superior temporal features, they still remained inapposite static detection mode. As a typical offline detection mode, Kang~\emph{et~al.} reported a TPN for tubelet proposal (i.e., temporally propagated boxes) so that multiple frames could be simultaneously processed to improve temporal consistency \cite{bib:TPN}. However, this batch-frame mode struggled to be qualified for real-world tasks. On the contrary, without the aid of any other temporal tools, we novelly develop a real-time online detection mode for videos using the idea of refinement. That is, refined anchors and refined feature sampling locations are generated with key frames, which would be temporally propagated for detection. Compared to most video detectors, our methods have a concise training process without the need for sequential images.

\subsection{Sampling for Detection}
It is widely accepted that spatial sampling is important to construct robust features. For example, Peng \emph{et al.} detected objects by an improved multi-stage particle window that can sample a small number of key features for detection \cite{bib:IPW}. In terms of CNN, canonical convolution is based on a square kernel that is not suited enough to variform objects. For augmenting the spatial sampling locations, Dai \emph{et al.} proposed deformable convolutional networks to combat fixed geometric structures in traditional convolution operation. The deformable convolution significantly boosted the detection accuracy of RFCN \cite{bib:DeformConv}. As for video detection, Bertasius \emph{et al.} used the deformable convolutions across time and constructed robust features for temporally describing objects \cite{bib:STSN}. Zhang \emph{et~al.} designed a feature consistency module with deformable convolution to reduce inconsistency in the single-stage pipeline \cite{bib:CRetina}. Wang \emph{et al.} proposed guided anchoring for RPN, Faster RCNN, and RetinaNet to achieve higher-quality region proposal \cite{bib:GA}. Creatively, we tend to capture accurate single-stage features, and more specifically, refined feature locations are produced based on refined anchors. Moreover, we propagate refinement information across time for video detection.

\section{Dual Refinement Network}
\label{sec:DRN}
In this section, the proposed DRNet will be presented. The network architecture is first briefed, then, we will demonstrate how to overcome two key single-stage drawbacks with anchor-offset detection. Next, our designed multi-deformable head is delineated, followed by the training and inference.

\subsection{Overall Architecture}

\subsubsection{Basic Structure}
As shown in Fig.~\ref{fig:DRN}, our proposed architecture is a single-shot network with a forward backbone for feature extraction. The network generates a fixed number of bounding boxes and corresponding classification scores, followed by the NMS for duplicate removal. Inheriting from RefineDet \cite{bib:RefineDet}, there is an anchor refinement module (ARM) and an object detection module (ODM) for two-step regression. ARM regresses coordinates for refined anchors, then feature offsets are predicted using anchor offsets. In ODM, a creative detection head is designed with deformable convolution for final classification and regression, whose inputs are ODM features, refined anchors, and feature offsets. Furthermore, a multi-deformable head is developed with multiple detection paths to leverage contextual information for detection.

\subsection{Anchor-Offset Detection}
\label{sec:AOD}
\subsubsection{From SSD to RefineDet, then to DRNet}
\begin{figure}[!t]
\centering
\subfigure[] { \label{fig:arch_comp}
\includegraphics[width=7.5cm]{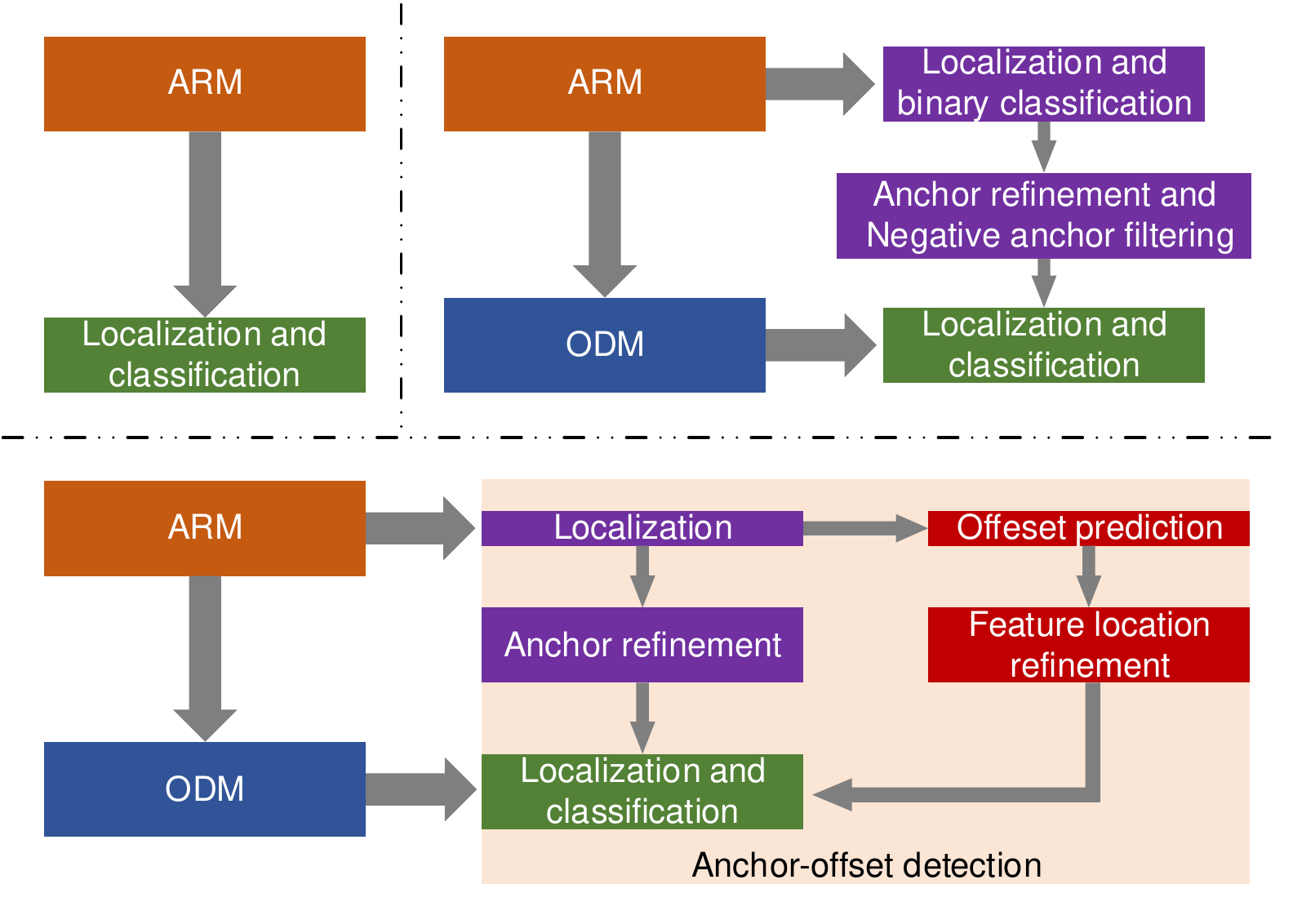}
}
\subfigure[] { \label{fig:aod}
\includegraphics[width=6.5cm]{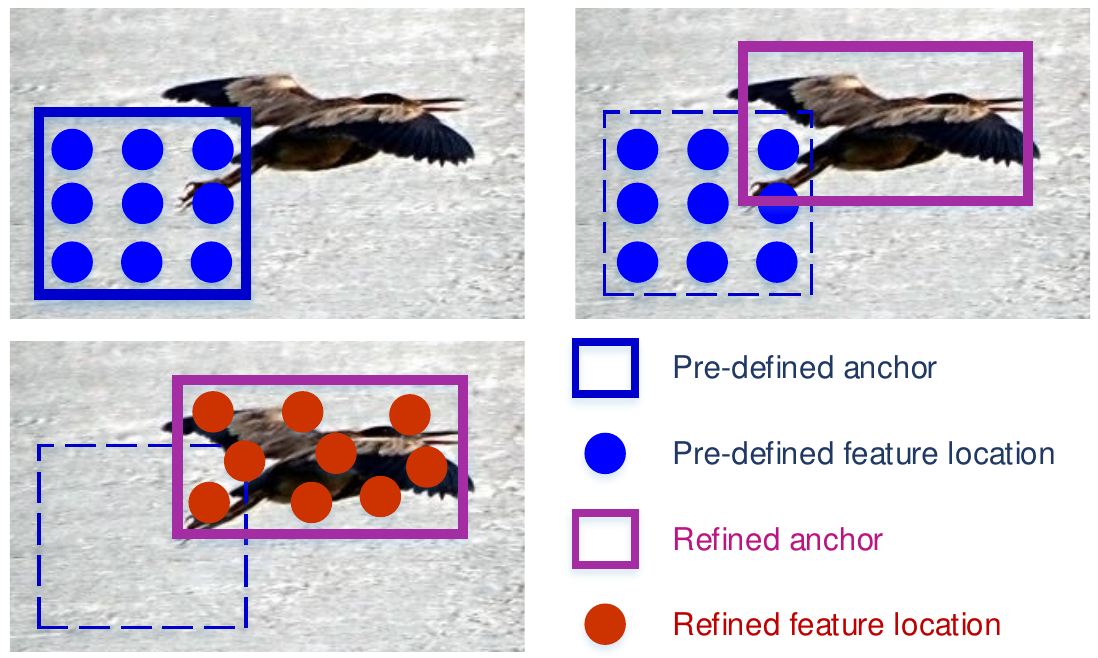}
}
\caption{Comparison of three single-stage detectors. (a) Structure sketch of SSD (left, top), RefineDet (right, top), and DRNet (bottom). (b) Detection modes of SSD (left, top), RefineDet (right, top), and DRNet (left, bottom). (b) shows the main idea of the anchor-offset detection.}
\label{fig:comp}
\end{figure}

As illustrated in Fig.~\ref{fig:arch_comp}, SSD directly detects objects with ARM features, whereas RefineDet adopts FPN for strong semantic information. Moreover, RefineDet develops an anchor refinement for more precision localization and a negative anchor filtering for addressing extreme class imbalance problem. In our DRNet, we inherit anchor refinement but discard the negative anchor filtering since training with hard negative mining \cite{bib:SSD} induces a similar effect. More specifically, a feature location refinement and a deformable detection head are proposed to combat another key drawback in the single-stage paradigm, i.e., inaccurate feature sampling locations.

In general, detection in traditional SSD-like manner is based on hand-crafted anchors which are rigid and usually inaccurate. Pre-defined anchors and fixed feature locations could not be suited enough to regress and classify objects (see the left top in Fig.~\ref{fig:aod}). Through preliminary localization, refined anchors in RefineDet are in favor of more precise coordinate prediction. However, RefineDet still uses inaccurate feature sampling locations (see the right top in Fig.~\ref{fig:aod}) for regression and classification. That is, the anchor refinement would incur serious anchor-feature misalignment. Thus, it is defective that the anchor refinement is leveraged alone. Overcoming these difficulties, our designed anchor-offset detection is able to achieve two-step regression and capture more accurate detection features in a single-stage pipeline (see the left bottom in Fig.~\ref{fig:aod}). This joint anchor-feature refinement manner is more reasonable than RefineDet.

\subsubsection{Anchor Refinement}
This process is analogous to RefineDet, i.e., ARM generates refined anchors that provide better initialization for the second-step regression. A location head performs convolution to generate anchor offset $ar$ using backbone-based ARM features $f_{ARM}$. That is, $ar=W_{ar}*f_{ARM}$, where $*$ denotes convolution ($W$ is the learnable convolutional weight). Note that $ar$ is the coordinate offset from original anchors. Anchor refinement is urgently necessary. On one hand, it highly relieves the difficulty of localization. On the other hand, it can guide feature location refinement.

\subsubsection{Deformable Detection Head}
According to deformable convolution \cite{bib:DeformConv}, a deformable detection head is designed to leverage the refinement information. The standard detection head in SSD uses a regular $3\times 3$ grid $\mathcal R$ to predict category probability and coordinates for a feature map cell. In the meantime, through careful anchor design, the respective field of $\mathcal R$ can describe a specific anchor region. Thus, the prediction can be given as $P_{p_0}=\sum_{p\in \mathcal R}w(p)\cdot f_{ODM}(p)$, where $P$ is the prediction of category probability or coordinate offset; $w$ is the convolution weight; $p$ represents positions in $\mathcal R$ while $p_0$ is the center; $f_{ODM}$ denotes ODM features.

However, the respective field of $\mathcal R$ usually fails to describe the refined anchor region (see the right top of Fig.~\ref{fig:aod}). Thereby, allowing $\mathcal R$ to deform to fit various anchor changes, the deformable detection head is developed to capture accurate features with the feature offset $\delta p$,
\begin{equation} \label{eqn:DDH}
P_{p_0}=\sum_{p\in \mathcal R}w(p)\cdot f_{ODM}(p+\delta p).
\end{equation}

\subsubsection{Feature Location Refinement}
The offset set $\Delta p=\{\delta p\}$ is computed with the input feature in original deform pipeline,
\begin{equation} \label{eqn:OriginalOffset}
\Delta p=W_{fr}*f_{ODM},
\end{equation}
where $W_{fr}$ is the convolutional weight. Nevertheless, there is a strong demand for describing the refined anchor regions with the deformed grids. Therefore, our feature offsets are predicted based on anchor offsets, i.e., feature location refinement,
\begin{equation} \label{eqn:FeatureRefine}
\Delta p=W_{fr}*ar.
\end{equation}

In detail, this operation is a convolution with $1\times 1$ kernels. Since each spatial element in $ar$ is coordinate offsets for refined anchors, its channel information is fused for feature location refinement. Note that anchor offsets and feature offsets are different in tensor shape. A man-made function can be designed to map anchor offsets to feature offsets, but we adopt a learnable mapping. Although the man-made manner can also promote refined feature locations to describe refined anchors, it still generates a conventional regular feature sampling region. Thus, the feature location refinement adopts a learnable manner to produce flexible feature sampling locations from multiple anchor offsets.

In this way, the refined feature locations can describe refined anchor regions more effectively. We call this detection mode anchor-offset detection, which can be formulated as
\begin{equation} \label{eqn:AOD}
\begin{array}l
P_{local}=(W_{local}*(f_{ODM}, \Delta p)) \oplus (ar \oplus ao)\\
P_{conf}=W_{conf}*(f_{ODM}, \Delta p).
\end{array}
\end{equation}
where $\oplus, ao$ represent anchor decoding operation \cite{bib:SSD} and the original anchor, respectively; $W*(f, \Delta p)$ denotes deformable convolution with $W$ as the weight. As $ar$ is the coordinate offset from $ao$, $ar \oplus ao$ is the refined anchor. The operation of two $\oplus$ is two-step regression that elevates the precision of localization, while $\Delta p$ is the feature offset that constructs the accurate single-stage detection features.

\subsection{Multi-Deformable Head}
CoupleNet developed local and global FCN to detect objects \cite{bib:CoupleNet}. The local FCN focused local features in a region proposal while the global one paid attention to the whole region-level features. In this way, more semantic information and underlying object relation are exploited for high-quality detection. Thus, taking aim at describing the object using original, shrunken, and expansile region-level features, a multi-deformable head is developed for the single-stage detector. The shrunken region-level features are in favor of leveraging local messages while the expansile region-level features contain more contextual information and object relation.

\begin{figure}[!t]
\centering
\includegraphics[width=7.5cm]{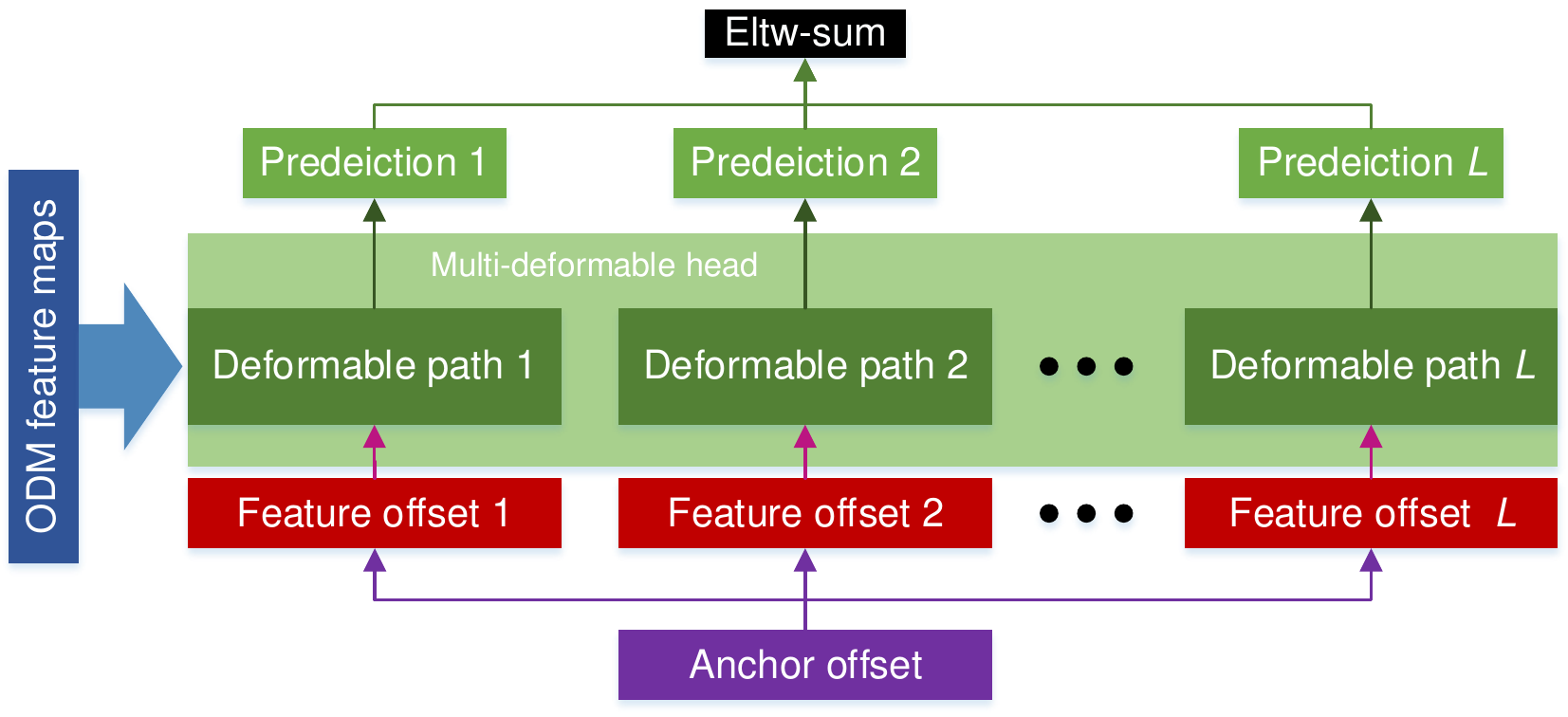}
\caption{Multi-deformable head. It is designed with different detection respective field. Multiple detection paths are induced, where feature location refinement is independent for each path. Their results are fused by summation.}
\label{fig:mdh}
\end{figure}

In this way, multiple detection head is designed with different respective field sizes, inducing multiple detection paths. As shown in Fig.~\ref{fig:mdh}, each of detection path is an anchor-offset detection, and their feature location refinement is independent. Besides, their results are fused with element-wise summation. The detection based on $L$ deformable paths can be given as
\begin{equation} \label{eqn:MDH}
\begin{array}l
P_{local}=(\sum_{l=1}^L W_{local_l}*(f_{ODM}, \Delta p_{l})) \oplus (ar \oplus ao) \\
P_{conf}=\sum_{l=1}^LW_{conf_l}*(f_{ODM}, \Delta p_{l}).
\end{array}
\end{equation}

\subsection{Training and Inference}
As for pre-defined anchor setting, each feature layer is associated with one specific scale of anchors, i.e., the anchor size of $[32,64,128,256]$ is adopted for $4$-scale feature maps from low-level to high-level, and $3$ anchors are tiled at each feature map cell with aspect ratios of $[1.0,2.0,0.5]$. In terms of optimization, an SGD optimizer with $0.9$ momentum and $0.0005$ weight decay is employed to train the whole network. Because of different sizes of datasets, the learning rate schedule is diverse for each dataset, which will be briefed latter.

A multi-task objective is designed to train DRNet including two localization losses $\mathcal L_{loc-ARM}, \mathcal L_{loc-ODM}$ and a confidence loss $\mathcal L_{conf}$, i.e., $\mathcal L = \frac{1}{N_{ARM}}\mathcal L_{loc-ARM} + \frac{1}{N_{ODM}}(\mathcal L_{loc-ODM}+\mathcal L_{conf})$, where $N$ is the number of positive boxes in ARM and ODM. $\mathcal L_{loc} = \sum_{i=1}^N sommothL1(p_i-g^*_i)$, where $g_i^*$ is the ground truth coordinates of the $i$-th positive anchor. Before computation of $\mathcal L_{loc}$, anchors should be determined to be positive or negative based on jaccard overlap \cite{bib:SSD}. We handle original anchors and refined anchors for $\mathcal L_{loc-ARM}$ and $\mathcal L_{loc-ODM}$, respectively, by the following processes. Firstly, each ground truth box is matched to anchors with the best jaccard overlap, then anchors with $>0.5$ overlap will be matched to corresponding ground truth box. Let $c^{cls}_i$ be the probability that the $i$-th predicted box belongs to class $cls$ ($cls=0$ for background). $\mathcal L_{conf} = -\sum_{i=1}^N \log(c^{cls}_i)-\sum_{i=1}^{\delta N}\log(c_k^0)$, where $\delta N$ negative anchors are selected by hard negative mining \cite{bib:SSD}. This operation selects a part of negative boxes with top loss values for training to address the problem with extreme foreground-background class imbalance, and $\delta=3$.

In inference phase, DRNet predicts confident object candidates (confident scores $>0.01$) in the manner of anchor-offset detection and multi-deformable head. Subsequently, these candidates are processed by NMS with $0.45$ jaccard overlap pre class and retain top $200$ (for COCO) or $300$ (for VOC and VID) high confident objects  as the final detections.

\section{Temporal Dual Refinement Networks}
\label{sec:TRN}
In this section, we present how to propagate refined anchors and refined feature sampling locations across time. Next, TRNet and TDRNet are formed. Then, we also describe the proposed soft refinement strategy.

\subsection{Architecture}
\begin{figure}[!t] \centering
\subfigure[] { \label{fig:TRN_arch}
\includegraphics[width=8cm]{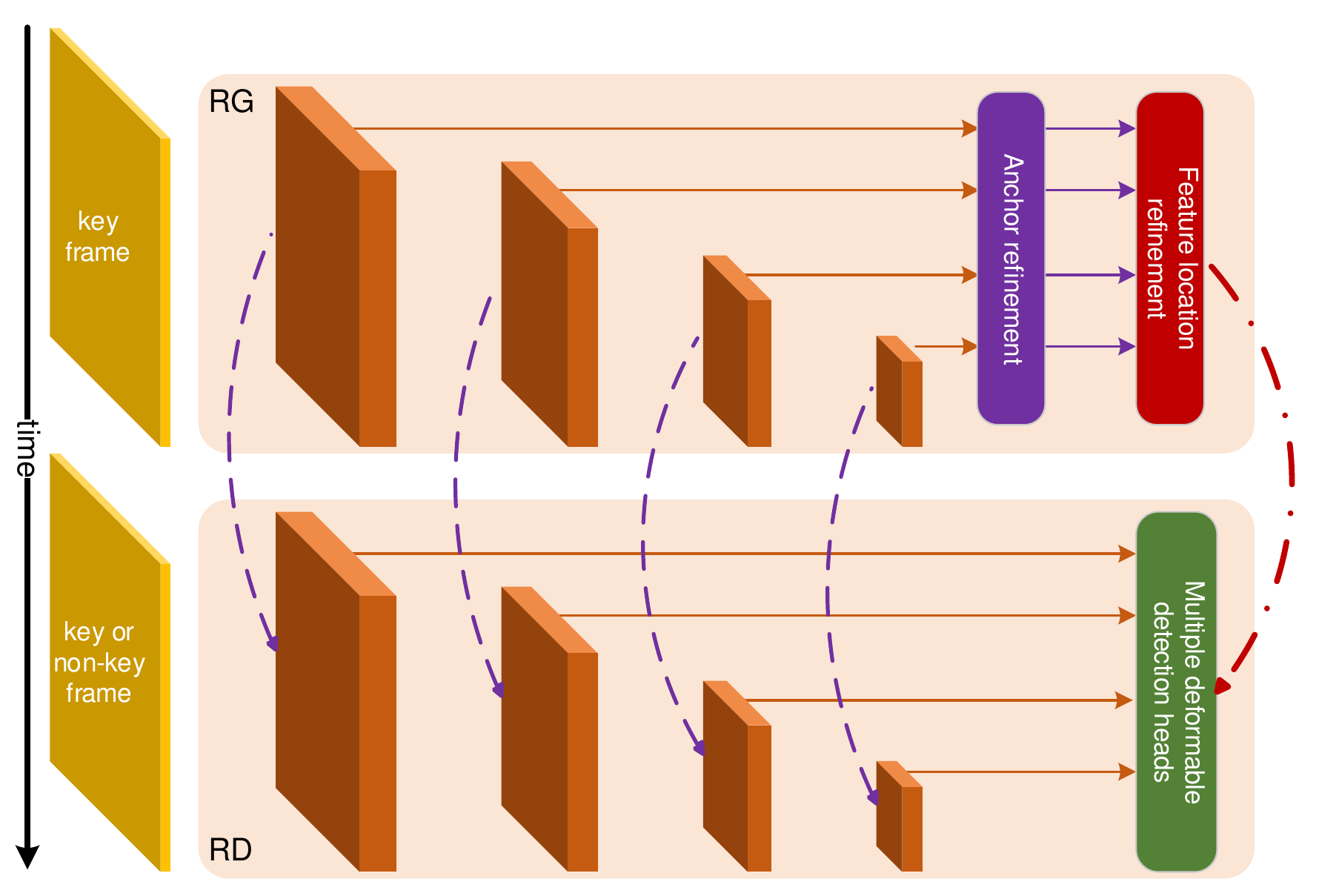}
}
\subfigure[] { \label{fig:TRN_train}
\includegraphics[width=8cm]{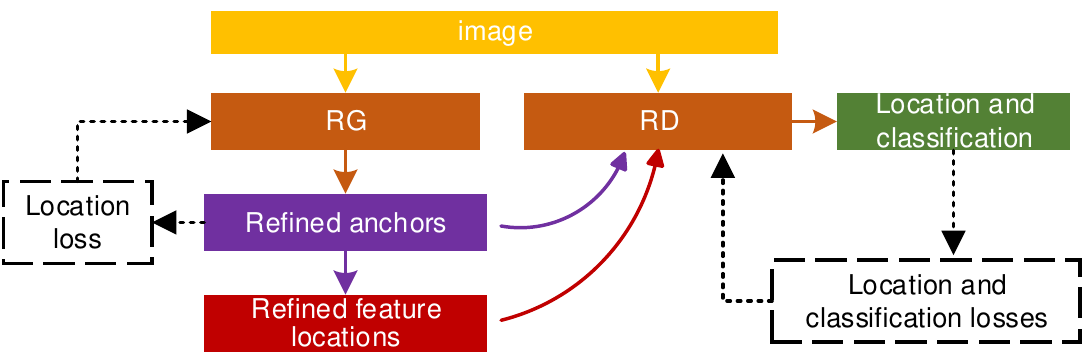}
}
\subfigure[] { \label{fig:TRN_test}
\includegraphics[width=8cm]{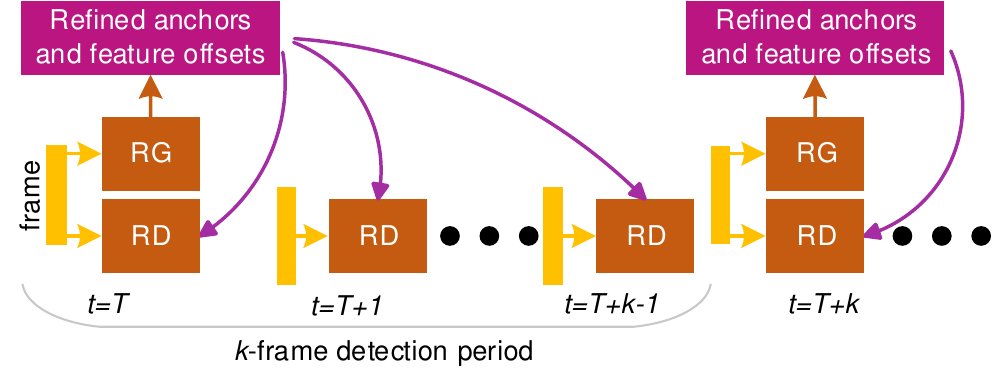}
}
\caption{Designs of the proposed TDRNet. (a) Network structure for RG and RD. RG generates refinement information while RD performs final detection. RG's outputs (i.e., anchor offsets and feature location offsets) serve as RD's inputs. (b) The training phase. (c) The testing phase.}
\label{fig:TRN}
\end{figure}

As shown in Fig.~\ref{fig:TRN_arch}, a reference generator (RG) and a refinement detector (RD) are designed in this section, both of which are constructed with similar structure, i.e., canonical SSD framework \cite{bib:SSD} with $4$-scale detection features. However, RG and RD have different training mode, parameters, and outputs. Like ARM in DRNet, RG predicts refinement information including refined anchors or both refine anchors and feature offsets. Similar to ODM in DRNet, RD takes over RG's outputs as references, and detect objects frame by frame. If RG only predicts refined anchors, the framework is called TRNet. When feature offsets are also predicted by RG, the anchor-offset detection with a deformable detection head is also employed by RD, and we call this structure TDRNet. That is, compared to TDRNet (see Fig.~\ref{fig:TRN_arch}), TRNet does not contain the module of feature location refinement, and its detection head is composed of traditional convolutions.

\subsection{Training}
In general, temporal detectors usually have a complex training process with sequential images. For example, TSSD developed a multi-step training strategy \cite{bib:TSSD}, and the initialization for multi-frame regression layer in TPN is complicated \cite{bib:TPN}. Conversely, the training process for TRNet and TDRNet is refreshingly concise, and we also eliminate the need of sequential training images. As shown in Fig.~\ref{fig:TRN_train}, during the training process, RG and RD play similar roles to DRNet's ARM and ODM, respectively. Thereby, both RG and RD can be trained with static images following DRNet's basic training settings and loss functions.

\subsection{Inference}
Consider a video as an image sequence, i.e., $V=\{I_0,I_1,...,I_M\}$. TRNet and TDRNet attempt to obtain frame-level detections $\{D_0,D_1,...,D_M\}$, where $D_m$ contains the boxes and class predictions of $I_m$. RG takes over $I_m$ and outputs anchor offset $ar$ and feature offset $\Delta p$,
\begin{equation} \label{eqn:RG}
ar_m, \Delta p_m=RG(I_m).
\end{equation}

Then, RD detects objects with $I_m$, $ar$, and $\Delta p$,
\begin{equation} \label{eqn:RD}
\begin{array}l
D_m=RD(I_m, ar, \Delta p) \\
= \left\{\begin{array}l
             (W_{local}*(f_{I_m}, \Delta p)) \oplus (ar \oplus ao)\\
              W_{conf}*(f_{I_m}, \Delta p),
             \end{array}
        \right.
\end{array}
\end{equation}
where $f_{I_m}$ is the feature extracted from $I_m$.

\begin{figure}[!t]
 \centering
\includegraphics[width=8cm]{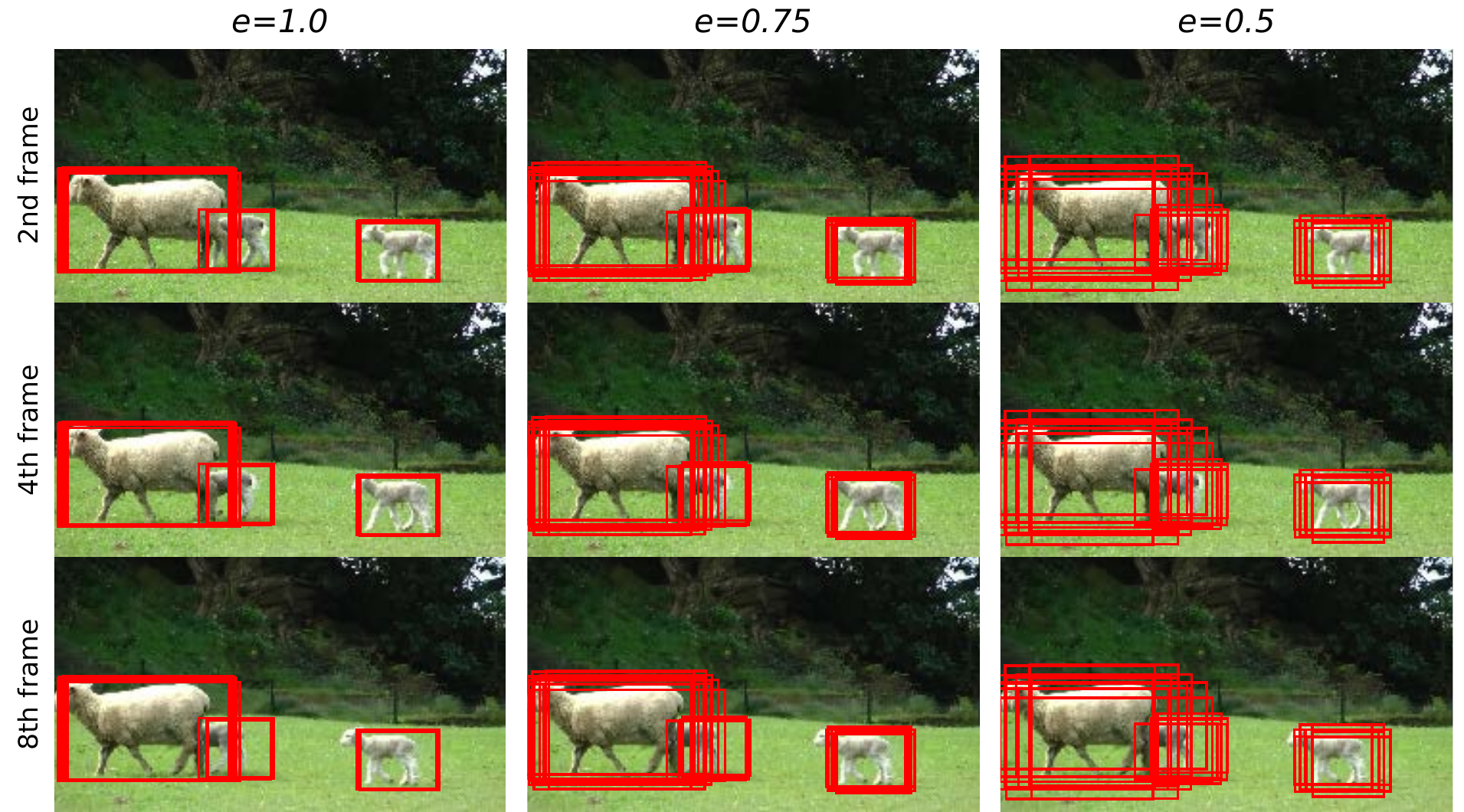}   
\caption{Soft refinement strategy. To match object motion with the previous refinement, a soft coefficient $e$ is introduced. Computed with the 1st frame, key refined anchors are demonstrated in the 2nd, 4th, and 8th frames. The decrease of $e$ reduces refinement intensity, producing loosely scattered refined anchors. That is, it is seen that refined anchors based on $e=1$ are compact. In contrast, refined anchors based on $e=0.75$ are more loosely scattered than that based on $e=1$, while refined anchors based on $e=0.5$ are more loosely scattered than that based on $e=0.75$.}
\label{fig:ref_anchor}
\end{figure}

\begin{figure*}[!t]
 \centering
\includegraphics[width=16cm]{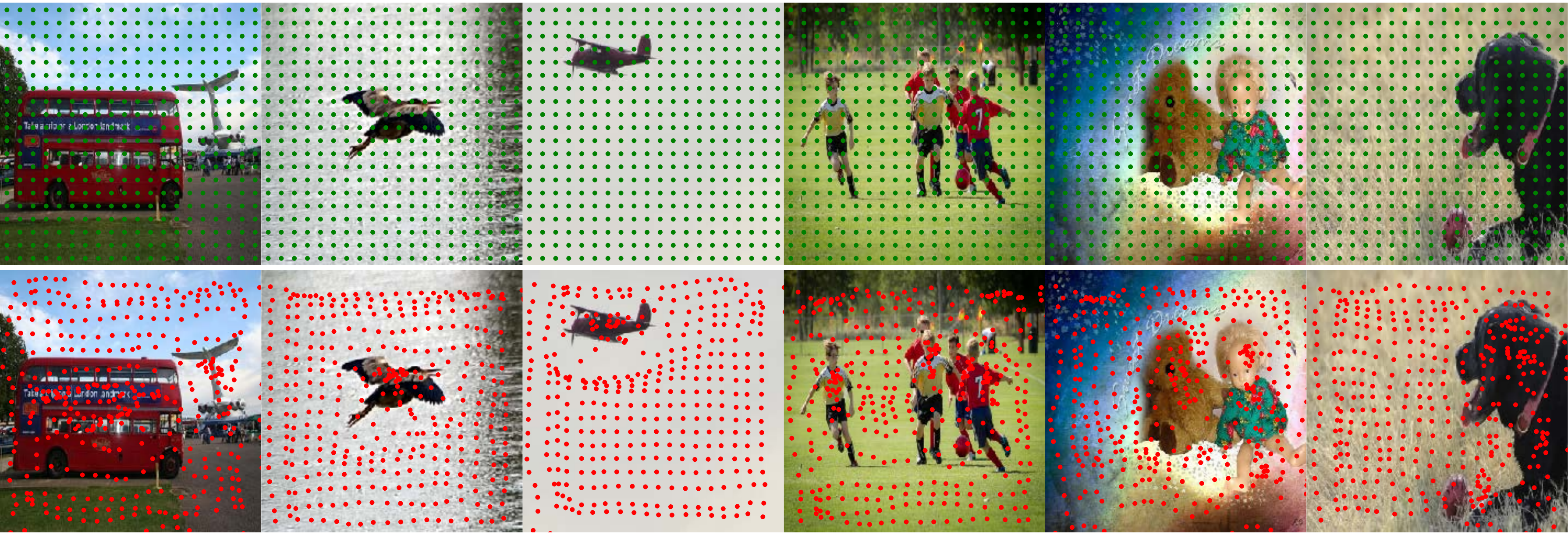}   
\caption{Visualization of refined feature sampling locations for \emph{Conv5\_3}. For better visualization, only the sampling centers (i.e., the center dot in left-bottom Fig~\ref{fig:aod}) are demonstrated. The original sampling centers are illustrated with green dots, which are regularly tiled on images. The red dots show the refined sampling centers that have a stronger capability of describing objects. These images are from VOC, COCO and ImageNet VID.}
\label{fig:feature_offset}
\end{figure*}

Despite the similar detection manner, it is apparent that RD is more computationally efficient than DRNet. Therefore, considering the temporal context in temporal vision, a key frame duration is used for RG to pursue a better trade-off between accuracy and speed. That is, only key frames will be processed by RG, while non-key frames are detected by RD with previous RG's outputs. Mathematically, in (\ref{eqn:RD}), $ar=ar_m,\Delta p=\Delta p_m$ for key frames, whereas $ar,\Delta p$ are from the previous key frame when detecting non-key frames. In this manner, $ar$ and $\Delta p$ are propagated as the temporal information. As illustrated in Fig.~\ref{fig:TRN_test}, using the first frame in a period, RG generates refinement references that will survive $k$ time stamps. Then, RD detects objects based on these references in the whole period. It is apparent that frequent reference update would lead to higher detection accuracy and more computational costs, so the trade-off between accuracy and speed can be adjusted by different $k$ setting.

Taking aim at adapting to various object motion, a soft refinement strategy is proposed with a soft coefficient $e$. In SSD, the intent of designing anchor is to use numerous boxes to cover the whole image as the prior knowledge, but significantly discarding anchor diversity, the refined anchors tend to surround foreground. Refined anchors are in favor of static detection, but objects in videos have a variety of motion properties or pose changes. Hence, the soft refinement strategy is designed to retain the anchor diversity for relatively long temporal detection period. The soft refinement can be given as $ar_{s}=ar \times e$, where $ar_{s}$ is the soft anchor offset, and as a scalar, $e\in [0,1]$ multiplies each element in a tensor. Because $ar$ is the offset from original anchors, $ar \times e$ can relax the intensity of anchor refinement by reducing offset magnitude so that refined anchors can be loosely scattered around objects. Referring to Fig.~\ref{fig:ref_anchor}, refined anchors are computed by the first frame in the period, then some key refined anchors are visualized in the $2$nd, $4$th, and $8$th frames. Without the soft refinement strategy, they gradually fail to be precisely aware of objects across time. For example, when $e=1$, refined anchors cannot surround the head of a sheep in the $8$th frame. This phenomenon causes regression difficulties for RD, prohibiting $k$ from increasing. That is, $e=1$ incurs that the refinement information can hardly be propagated in a relatively long range of time series. When $e=0.75$ or $0.5$, this drawback is mitigated so that the detection period can be longer for a better trade-off between accuracy and speed.

\section{Experiments and Discussion}
\label{sec:EXP}
Our methods are implemented under the Pytorch framework. The training and experiments are carried out on a workstation with an Intel 2.20 GHz Xeon(R) E5-2630 CPU, NVIDIA TITAN-1080 GPUs, CUDA 8.0, and cuDNN v7. Our approaches are trained and evaluated on PASCAL VOC \cite{bib:VOC}, COCO \cite{bib:COCO}, and ImageNet VID \cite{bib:ImageNet} datasets. Furthermore, we applied TDRNet to online underwater object detection.

\subsection{Ablation Studies of DRNet320-VGG16 on VOC 2007}
Experiments on PASCAL VOC 2007 are first conducted with VGG16 \cite{bib:VGG} as the backbone to study the proposed dual refinement mechanism in detail. In this section, the models are trained on the union set of VOC 2007 trainval and VOC 2012 trainval ($16,551$ images, denoted as ``07+12''), and evaluated on VOC 2007 test set ($4,952$ images). We use mAP to describe the detection accuracy. For the convenience of comparison, RefineDet without negative anchor filtering is employed as the baseline, whose mAP is $79.1\%$ in our re-produced Pytorch implementation (Note that it is $79.5\%$ in original Caffe implementation). The changes of mAP caused by various model designs are shown in Table~\ref{tab:mAP}.

\begin{table}[!t]
\renewcommand{\arraystretch}{1.1}
\setlength\tabcolsep{5.2pt}
\caption{{\bf Ablation Studies of DRNet320 on VOC 2007.} The baseline is $79.1\%$}
\label{tab:mAP}
\centering
\begin{tabular}{c | c | c c c c c}
\Xhline{1.5pt}
Component                    &   & \multicolumn{5}{c}{DRNet320-VGG16}     \\
\hline
multi-deformable head?       &            &           &\checkmark  &             &              &\checkmark \\
feature location refinement?   &            &\checkmark &\checkmark  &             &\checkmark    &\checkmark \\
deformable detection head?   &\checkmark  &\checkmark &\checkmark  &             &\checkmark    &\checkmark \\
BN for VGG\&extra?           &            &           &            &\checkmark   &\checkmark    &\checkmark \\
\hline
mAP(\%)                      & $78.3$     & $79.8$    &$80.5$      &  $81.1$     & $81.7$       & 82.0 \\
\Xhline{1.5pt}
\end{tabular}
\end{table}

\begin{table}[!t]
\renewcommand{\arraystretch}{1.1}
\setlength\tabcolsep{8pt}
\caption{{\bf Effectiveness of various multi-deformable head designs.} A variety of detection paths with different convolutional kernel size ($ks$) and dilation ($di$) are used to validate the efficacy our designs.}
\label{tab:MHD}
\centering
\begin{tabular}{c | c c c c c}
\Xhline{1.5pt}
$ks=5\times 5, di=1$?            &            &            &           &\checkmark  &\checkmark\\
$ks=3\times 3,di=2$?             &            &            &\checkmark &            &          \\
$ks=1\times 1,di=1$?             &            &\checkmark  &\checkmark &\checkmark  &          \\
$ks=3\times 3,di=1$?             & \checkmark &\checkmark  &\checkmark &\checkmark  &\checkmark\\
\hline
mAP(\%)                        &  $79.8$     & $79.8$     & $79.4$     &$80.5$     & $80.3 $ \\
\Xhline{1.5pt}
\end{tabular}
\end{table}

\begin{table*}[!t]
\renewcommand{\arraystretch}{1.1}
\setlength\tabcolsep{10pt}
\caption{{\bf Results on Pascal VOC 2007 and 2012 test dataset}. ``Train data'' is used for VOC 2007 training, and that of VOC 2012 contains an extra VOC 2007 test set. ``S'' denotes instance segmentation labels. ``+'' indicates multi-scale testing. Bold fonts indicate the best.}
\label{tab:VOC}
\centering
\begin{tabular}{c | c| c | c | c | c | c | c}
\Xhline{1.5pt}
\multirow{2}{0.8cm}{Method}    & \multirow{2}{1.1cm}{Backbone}  & \multirow{2}{1.1cm}{Train~data}     & \multirow{2}{1.0cm}{Input~size}  & \multirow{2}{0.8cm}{\#Boxes} &  \multirow{2}{0.5cm}{FPS}       & \multicolumn{2}{c}{mAP} \\
 \cline{7-8}
&&&&&& VOC 2007    & VOC 2012  \\
\hline
\emph{two-stage}                  &           & &                      &                 &                  &          &     \\
Faster RCNN \cite{bib:FasterRCNN} & VGG16    &07+12 & $1000\times 600$     & $300$           &  $7$             & $73.2$   &$70.4$        \\
Faster RCNN \cite{bib:FasterRCNN} & ResNet101&07+12 & $1000\times 600$     & $300$           &  $2.4$           & $76.4$   &$73.8$        \\
RFCN \cite{bib:RFCN}             & ResNet101&07+12 & $1000\times 600$     & $300$           &  $9$             & $80.5$   &$77.6$        \\
CoupleNet \cite{bib:CoupleNet}    & ResNet101&07+12 & $1000\times 600$     & $300$           &  $8.2$           & $82.7$ &$80.4$        \\
Attention CoupleNet \cite{bib:ACoupleNet}& ResNet101 &07+12+S & $1000\times 600$ & $300$     &  $6.9$           & $83.1$ &$81.0$        \\
\hline
\hline
\emph{single-stage}                 &                             &                      &       &                &         &    \\
YOLO \cite{bib:YOLO}             & GoogleNet \cite{bib:GoogleNet}&07+12 & $448\times 448$      & 98    &  $45.0$          & $63.4$ & $57.9$      \\
SSD321 \cite{bib:DSSD}           & ResNet101                    &07+12 & $321\times 321$      & 17080 &  $11.2$        & $77.1$ & $75.4$        \\
SSD300 \cite{bib:SSD}            & VGG16                        &07+12 & $300\times 300$      & 8732  &  $120.0^*$          & $77.2$ & $75.8$        \\
DSOD300 \cite{bib:DSOD}          & DenseNet \cite{bib:DenseNet}  &07+12 & $300\times 300$      & 8732  &  $17.4$        & $77.7$ & $76.3$    \\
YOLOv2 \cite{bib:YOLOv2}         & Darknet19                    &07+12 & $544\times 544$      & 1445  &  $40.0$          & $78.6$ & $73.4$      \\
DSSD321 \cite{bib:DSSD}          & ResNet101                    &07+12 & $321\times 321$      & 17080 &  $9.5$         & $78.6$ & $76.3$        \\
SSD512 \cite{bib:SSD}            & VGG16                        &07+12 & $512\times 512$      & 24564 &  $34.7^*$          & $79.8$ & $78.5$        \\
RefineDet320 \cite{bib:RefineDet}& VGG16                        &07+12 & $320\times 320$      & 6375  &  $60.0^*$        & $80.0$ & $78.1$        \\
RFBNet300 \cite{bib:RFB}         & VGG16                        &07+12 & $300\times 300$      & 8808  &  $83.0^*$          & $80.5$ & -             \\
SSD513 \cite{bib:DSSD}           & ResNet101                    &07+12 & $513\times 513$      & 43688 &  $6.8$         & $80.6$ & $79.4$        \\
DSSD513 \cite{bib:DSSD}          & ResNet101                    &07+12 & $513\times 513$      & 43688 &  $5.5$         & $81.5$ & $80.0$        \\
RefineDet512 \cite{bib:RefineDet}& VGG16                        &07+12 & $512\times 512$      & 16320 &  $24.1$        & $81.8$ & $80.1$        \\
RFBNet512 \cite{bib:RFB}         & VGG16                        &07+12 & $512\times 512$      & 24692 &  $38.0^*$          & $82.2$ & -      \\
\hline
DRNet320                & VGG16                 &07+12        & $320\times 320$         &6375 &  $55.2^*$        & $82.0$ & $79.3^\dagger$        \\
DRNet512                & VGG16                 &07+12        & $512\times 512$         &16320&  $32.2^*$        & $ 82.8$ & $ 80.6^\dagger$        \\
DRNet320+               & VGG16                 &07+12        & $320\times 320$         &6375 &  $-$        & $83.9$ & $83.1^\dagger$        \\
DRNet512+               & VGG16                 &07+12        & $512\times 512$         &16320&  $-$        & $\bf 84.4$ & $\bf 83.6^\dagger$        \\
\Xhline{1.5pt}
\end{tabular}
\begin{flushleft}
\begin{footnotesize}
$\dagger$: in VOC 2012 test server: \url{http://host.robots.ox.ac.uk:8080/anonymous/18COCB.html},\quad
\url{http://host.robots.ox.ac.uk:8080/anonymous/V1DWET.html}, \quad
\url{http://host.robots.ox.ac.uk:8080/anonymous/KCHPYZ.html}, \quad
\url{http://host.robots.ox.ac.uk:8080/anonymous/SZHWN4.html}. *: Pytorch speed.
\end{footnotesize}
\end{flushleft}
\end{table*}

\subsubsection{Anchor-Offset Detection}
The anchor-offset detection contains an anchor refinement, a feature location refinement, and a deformable detection head, the first of which has been studied by \cite{bib:RefineDet}, so we focus on the latter two components. At first, the deformable detection head without feature location refinement is tested. Following \cite{bib:DeformConv}, the offsets are computed with ODM features (referring to (\ref{eqn:OriginalOffset})). As a result, this change leads to a $0.8\%$ mAP drop. In our opinion, this should be attributed to improper offsets. That is, the refined anchors are computed with ARM while the feature offsets are from ODM, so they are independent, making refined feature locations still fail to describe refined anchor regions.

The refined anchors have been displayed in Fig.~\ref{fig:ref_anchor}, so the refined feature sampling locations are also demonstrated in Fig.~\ref{fig:feature_offset} to better explain the advantages of the proposed anchor-offset detection. For better visualization, only the sampling centers (i.e., the center dot in left-bottom Fig.~\ref{fig:aod}) are demonstrated. Referring to green dots in Fig.~\ref{fig:feature_offset}, the pre-defined detection features are regularly fixed on feature maps (their locations are mapped to the original images for visualization). This design is justified for the traditional SSD since anchors are also tiled in the same manner. However, the refined anchors tend to surround objects for more precision localization (see Fig.~\ref{fig:ref_anchor}), so it is reasonable that the feature locations should have the same tendency. As shown with red dots, gathering towards objects, the refined feature locations are more suitable for regression and classification. Moreover, in some areas away from objects, the refined feature locations would not blindly shift towards targets so that the detection capability for the whole image can be maintained.

Therefore, the operation of the proposed feature location refinement is crucial to capture accurate detection features. Following the pipeline of anchor-offset detection, the refined feature locations are tightly associated with refined anchors. Thus, a $0.7\%$ mAP rise (i.e., $79.8\%$ vs. $79.1\%$) is induced.

\subsubsection{Multi-Deformable Head}
For leveraging more contextual information for detection, multiple detection paths are devised with various respective field sizes, or convolution kernel size and dilation. The effectiveness of various multi-deformable designs is shown in Table~\ref{tab:MHD}. At first, the $1\times 1$ grid is employed to utilize shrunken region-level features, but it incurs negligible effectiveness. The $1\times 1$ grid should have focused on most suitable local parts for detection, but feature offsets are computed with anchor offsets in our pipeline, ignoring suitable local parts. Then, the $3\times 3$ grid with dilation is devised as one of the detection paths, but it leads to a $0.4\%$ drop in mAP. Although it expands the respective field, the dilated $3\times 3$ grid splits features and fails to describe objects effectively. Covering this shortage, the $5\times 5$ grid without dilation works more effectively, and it invites a $0.7\%$ mAP rise (i.e., $80.5\%$ vs. $79.8\%$) since more contextual information is involved. Moreover, the $1\times 1$ detection path is removed, and this more efficient design still can reach $80.3\%$ in mAP. These comparisons also indicate that the improvement comes from above-analyzed reasons rather than increasing parameter size.

\subsubsection{Towards More Effective Training}
Batch normalization \cite{bib:BN} is introduced to the backbone for more effective training, and a significant improvement in accuracy is incurred, i.e., $81.1\%$ mAP. Then, the anchor-offset detection and multi-deformable head further boost the performance. Referring to Table~\ref{tab:mAP}, removing the multi-deformable head leads to a $0.3\%$ drop in mAP, and removing the anchor-offset detection invites another $0.6\%$ mAP drop. Thus, our designs are still efficient, making a superior detection performance with such a small input image, i.e., $82.0\%$ mAP and $320\times 320$ input size.

\subsection{Results on VOC 2007}
We use the initial learning rate of $0.001$ for the first $130$ training epochs, then use the learning rate of $0.0001$ for the next $40$ epochs and $0.00001$ for another $40$ epochs. Referring to Table~\ref{tab:VOC}, our DRNet320 achieves $82.0\%$ mAP surpassing all methods with such small inputs by a large margin. When compared to SSD300, our method outperforms it by $4.8$ points (i.e., $82.0\%$ vs. $77.2\%$), and DRNet320 further improves mAP by $2.0\%$ as for RefineDet320 (i.e., $82.0\%$ vs. $80.0\%$). Compared to RFBNet300, our DRNet320 also has $1.5$-point higher mAP (i.e., $82.0\%$ vs. $80.5\%$).

For $512\times 512$ input size, DRNet512 obtains $82.8\%$ mAP that is also competitive with all compared methods. Only Attention CoupleNet \cite{bib:ACoupleNet} has slightly higher mAP than ours (i.e., $82.8\%$ vs. $83.1\%$). However, Attention CoupleNet uses ResNet101 \cite{bib:ResNet} as its backbone, and its results come with larger input size (i.e., $1000\times 600$). Besides, Attention CoupleNet introduces extra segmentation annotations to its multi-scale training processing. In addition, DRNet512's inference speed surpasses that of Attention CoupleNet by a large margin (i.e., $32.2$ vs. $6.9$ FPS). Therefore, the proposed DRNet achieves a better trade-off between accuracy and speed. To relieve the impact of relatively small input size, we leverage multi-scale strategy for testing, and DRNet320 and DRNet512 can obtain $83.9\%$, $84.4\%$ mAP, respectively.

\subsection{Results on VOC 2012}
More challenging VOC 2012 dataset is employed to evaluate our proposed designs, and we use the union set of VOC 2007 and VOC 2012 trainval sets plus VOC 2007 test set ($21,503$ images) for training in this experiment, and test models on VOC 2012 test set ($10,991$ images). The learning rate schedule is consistent with VOC 2007 training. Referring to Table~\ref{tab:VOC}, our DRNet320 obtains $79.3\%$ mAP that outmatch all compared methods with similar small input sizes. With $512\times 512$ input size, DRNet512 improves the mAP to $80.6\%$, which validates the effectiveness of our designs once again. Additionally, with multi-scale testing, $83.1\%$ and $83.6\%$ mAP are induced by DRNet320 and DRNet512.

\begin{table*}[!t]
\renewcommand{\arraystretch}{1.1}
\setlength\tabcolsep{5pt}
\caption{{\bf Results on COCO test-dev}. ``AP'' is evaluated at IoU thresholds from $0.5$ to $0.95$. ``AP@0.5'': PASCAL-type metric, IoU$=0.5$. ``AP@0.75'': evaluated at IoU$=0.75$. AP$_S$, AP$_M$ , AP$_L$: AP at different scales. ``+'' indicates multi-scale testing. Bold fonts indicate that we draw readers' attention to a deep comparison.}
\label{tab:COCO}
\centering
\begin{tabular}{c | c| c | c c c | c c c | c }
\Xhline{1.5pt}
Method    & Backbone  & Train~data   &  AP   & AP@0.5  &AP@0.75 &  AP$_S$ & AP$_M$  &  AP$_L$ & Time \\
\hline
\emph{anchor-free}                 &           & &                      &                 &                  &          &     & &\\
ExtremeNet \cite{bib:ExtremeNet}   & Hourglass104     &trainval35k     & $40.2$ & $55.5$ & $43.2$ & $20.4$ & $43.2$ & $53.1$ &$-$\\
CornerNet \cite{bib:CornerNet}     & Hourglass104     &trainval35k     & $40.6$ & $56.4$ & $43.2$ & $19.1$ & $42.8$ & $54.3$  &$-$   \\
FCOS \cite{bib:FCOS}     & ResNet101        &trainval35k     & $41.5$ & $60.7$ & $45.0$ & $24.4$ & $44.8$ & $51.6$ &$-$\\
CenterNet \cite{bib:CenterNet}     & Hourglass104     &trainval35k     & $42.1$ & $61.1$ & $45.9$ & $24.1$ & $45.5$ & $52.8$ &$-$\\
\hline
\hline
\emph{anchor-based two-stage}                  &           & &               &                 &                  &          &     & &\\
Faster RCNN$^\dagger$ \cite{bib:FasterRCNN} & MobileNet &trainval32k  & $19.8$     &$-$& $-$ & $-$ & $-$ & $-$       & $-$\\
Faster RCNN \cite{bib:FasterRCNN}     & VGG16    &train     & $24.2$     & $45.3$ & $23.5$ & $7.7$ & $26.4$ & $37.1$ & $147$ ms\\
Faster RCNN \cite{bib:FasterRCNN} & ResNet101    &trainval & $29.4$ & $48.0$ & $-$ & $9.0$ & $30.5$ & $47.1$       & $-$\\
RFCN \cite{bib:RFCN}             & ResNet101&trainval  & $29.9$     & $51.9$     &   $-$       & $10.8$  & $32.8$   & $45.0$    & $110$ ms  \\
{\bf Deformable Faster RCNN} \cite{bib:DeformConv} &ResNet101 &trainval & $\bf33.1$ & $\bf50.3$ & $-$ & $\bf11.6$ & $\bf34.9$ & $\bf51.2$  & $-$\\
{\bf CoupleNet} \cite{bib:CoupleNet}    & ResNet101&trainval  & $\bf34.4$   &$\bf 54.8$  &$\bf 37.2$  &$\bf13.4$  &$\bf38.1$ &$\bf50.8$  & $-$    \\
{\bf Faster RCNN+++} \cite{bib:ResNet}&ResNet101-c4&trainval &$\bf34.9$ & $\bf55.7$ &$\bf37.4$ & $\bf15.6$ & $\bf38.7$ & $\bf50.9$ & $3.36$ s\\
{\bf Deformable RFCN} \cite{bib:DeformConv} &ResNet101 &trainval & $\bf34.5$ & $\bf55.0$ & $-$ & $\bf14.0$ & $\bf37.7$ & $\bf50.3$ & $125$ ms \\
{\bf Attention CoupleNet} \cite{bib:ACoupleNet}& ResNet101   &trainval+S &$\bf35.4$ & $\bf55.7$ &$\bf37.6$ &$\bf13.2$ &$\bf38.6$ &$\bf52.5$  & $-$   \\
Faster RCNN w/ FPN \cite{bib:FPN}       &ResNet101 &trainval35k &$36.$2 & $59.1$ & $39.0$ & $18.2$ & $39.0$ & $48.2$& $240$ ms \\
\hline
\hline
\emph{anchor-based single-stage}              &                               &            &         &        &        &       &        & &\\
SSD300$^\dagger$ \cite{bib:SSD}  & MobileNet                     &trainval35k & $19.3$  & $-$      & $-$      & $-$     & $-$      &  $-$ & $-$\\
RFBNet300$^\dagger$ \cite{bib:RFB}& MobileNet                    &trainval35k & $20.7$  & $-$      & $-$      & $-$     & $-$      &  $-$  & $-$\\
YOLOv2 \cite{bib:YOLOv2}         & Darknet19                    &trainval35k & $21.6$  & $44.0$ & $19.2$ & $5.0$ & $22.4$ & $35.5$  & $25$ ms\\
SSD300 \cite{bib:SSD}            & VGG16                        &trainval35k & $25.1$  & $43.1$ & $25.8$ & $6.6$ & $25.9$ & $41.4$  & $12$ ms\\
DSSD321 \cite{bib:DSSD}          & ResNet101                    &trainval35k & $28.0$  & $46.1$ & $29.2$ & $7.4$ & $28.1$ & $47.6$    & $-$\\
SSD512 \cite{bib:SSD}            & VGG16                        &trainval35k & $28.8$  & $48.5$ & $30.3$ & $10.9$& $31.8$ & $43.5$    & $28$ ms  \\
RefineDet320 \cite{bib:RefineDet}& VGG16                        &trainval35k & $29.4$  & $49.2$ & $31.3$ & $10.0$& $32.0$ & $44.4$ & $-$ \\
RFBNet300 \cite{bib:RFB}         & VGG16                        &trainval35k & $30.3$  & $49.3$ & $31.8$ & $11.8$& $31.9$ & $45.9$ & $15$ ms \\
RefineDet320 \cite{bib:RefineDet}& ResNet101                    &trainval35k & $32.0$  & $51.4$ & $34.2$ & $10.5$& $34.7$ & $50.4$ & $-$ \\
SSD513 \cite{bib:DSSD}           & ResNet101                    &trainval35k & $31.2$  & $50.4$ & $33.3$ & $10.2$& $34.5$ & $49.8$ & $-$ \\
YOLOv3-608 \cite{bib:YOLOv3}           & DarkNet53              &trainval35k & $33.0$  & $57.9$ & $34.4$ & $18.3$& $35.4$ & $41.9$ & $51$ ms\\
{\bf RefineDet512} \cite{bib:RefineDet}& VGG16               &trainval35k & $\bf33.0$  & $\bf54.5$ & $\bf35.5$ & $\bf16.3$& $\bf36.3$ &$\bf44.3$ & $-$ \\
{\bf DSSD513} \cite{bib:DSSD}          & ResNet101           &trainval35k & $\bf33.2$  & $\bf53.3$ & $\bf35.2$ & $\bf13.0$& $\bf35.4$ &$\bf51.1$ &$182$ ms\\
{\bf RetinaNet500} \cite{bib:RetinaNet}& ResNet101           &trainval35k & $\bf34.4$  & $\bf53.1$ & $\bf36.8$ & $\bf14.7$& $\bf38.5$ &$\bf49.1$ & $90$ ms\\
{\bf RFBNet512} \cite{bib:RFB}         & VGG16               &trainval35k & $\bf34.4$  & $\bf55.7$ & $\bf36.4$ & $\bf17.6$& $\bf37.0$ &$\bf47.6$ & $33$ ms\\
RefineDet512 \cite{bib:RefineDet}& ResNet101                 &trainval35k & $36.4$  & $57.5$    & $39.5$    & $16.6$ & $39.9$ & $51.4$ & $-$\\
GA-RetinaNet \cite{bib:GA}& ResNet50                  &train       & $37.1$  & $56.9$ & $40.0$ & $20.1$ & $40.1$ & $48.0$ & $-$\\
Cas-RetinaNet800 \cite{bib:CRetina}  & ResNet101      &trainval35k & $41.1$  & $60.7$ & $45.0$ & $23.7$ & $44.4$ & $52.9$ & $-$ \\
\hline
DRNet320$^\dagger$      & MobileNet          &trainval35k      & $26.0/25.7$  & $45.3$ & $26.8$ & $8.0$& $28.7$ & $38.9$    & $27$ ms \\
DRNet512$^\dagger$      & MobileNet          &trainval35k      & $28.5/28.4$  & $49.8$ & $29.6$ & $14.3$& $32.1$ & $36.6$   & $51$ ms \\
DRNet320                & VGG16             &trainval35k      & $30.5$     & $51.2$ & $32.3$ & $11.2$& $33.9$ & $44.9$      & $29$ ms \\
\bf DRNet512            & VGG16             &trainval35k      & $\bf34.3$  & $\bf57.1$ & $\bf36.4$ & $\bf17.9$& $\bf38.1$ & $\bf44.8$     & $53$ ms   \\
DRNet320                & RedNet101         &trainval35k      & $33.5$     &$53.4$ & $35.9$ & $11.5$& $37.4$ & $50.6$     & $36$ ms \\
DRNet512                & ResNet101         &trainval35k      & $38.6$     &$60.3$  & $42.2$ & $19.0$& $43.2$ & $52.7$    & $61$ ms \\
DRNet320+               & VGG16             &trainval35k      & $35.4$     &$57.8$ & $37.7$ & $19.7$& $38.5$ & $45.7$     & $-$ \\
DRNet512+               & VGG16             &trainval35k      & $37.9$     &$61.6$ & $40.3$ & $22.6$& $40.4$ & $48.4$     & $-$ \\
DRNet320+               & RedNet101         &trainval35k      & $39.2$     &$61.3$ & $42.3$ & $21.4$& $42.8$ & $51.4$     & $-$ \\
DRNet512+               & ResNet101         &trainval35k      & $42.4$     &$65.5$  & $46.1$ & $25.7$& $45.3$ & $55.0$ & $-$ \\
\Xhline{1.5pt}
\end{tabular}
\begin{flushleft}
\begin{footnotesize}
$\dagger$: Prior MobileNet-based models are tested on COCO minival2014, so our MobileNet-based AP is reported as ``test-dev/minival2014''.
\end{footnotesize}
\end{flushleft}
\end{table*}

\begin{figure}[!t]
\centering
\subfigure { \label{fig:error_person}
\includegraphics[width=4cm]{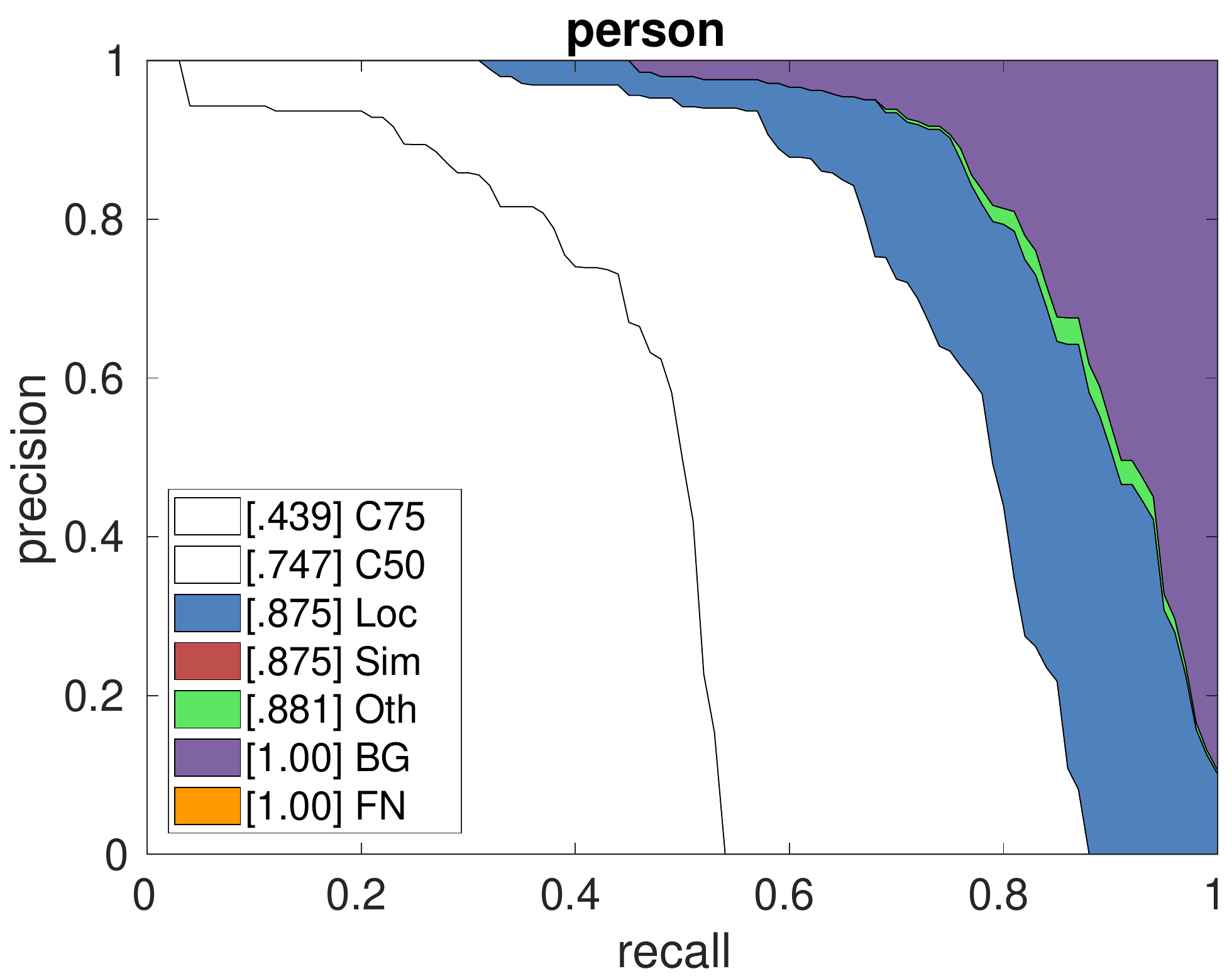}
}
\subfigure { \label{fig:error_vehicle}
\includegraphics[width=4cm]{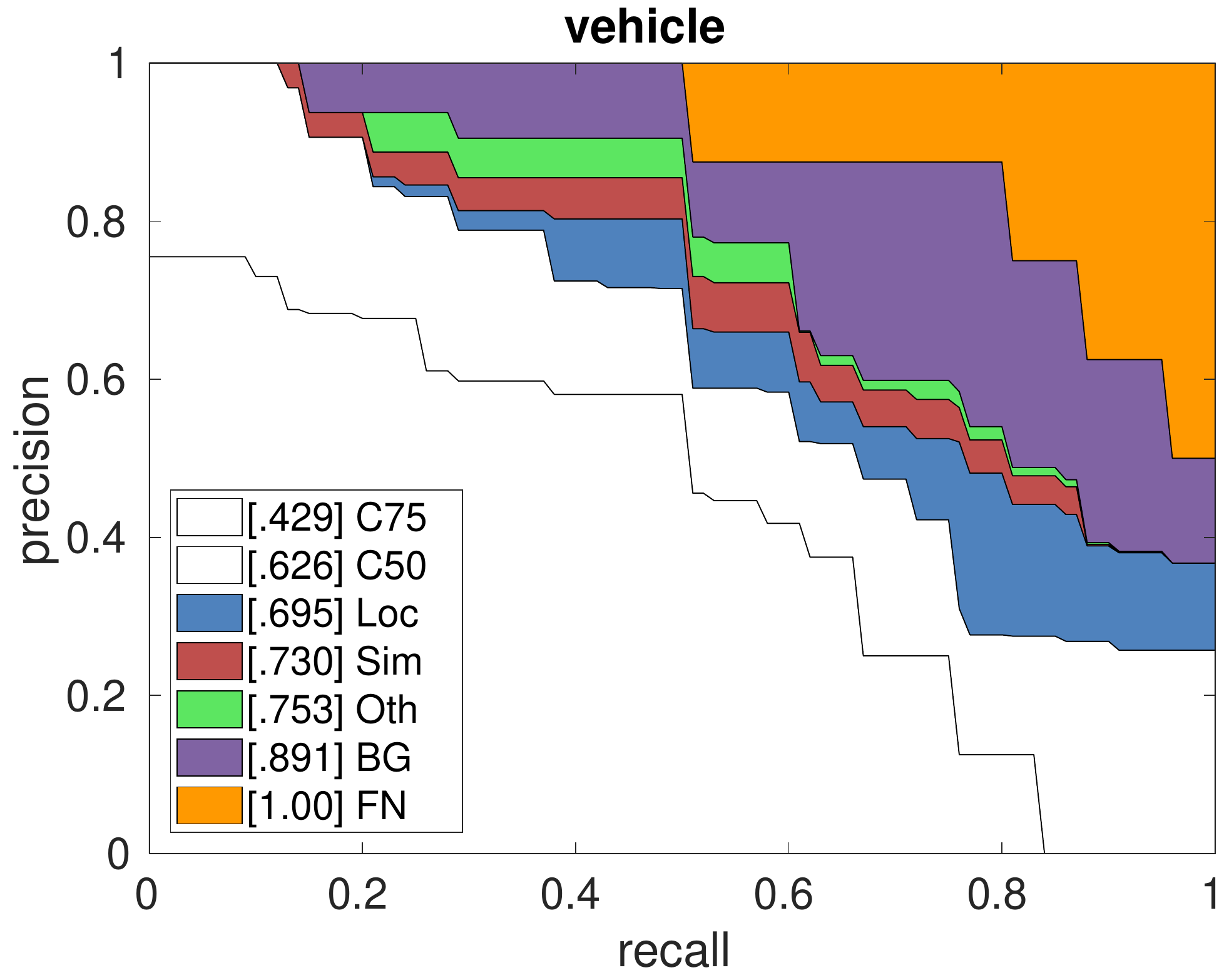}
}
\subfigure { \label{fig:error_furniture}
\includegraphics[width=4cm]{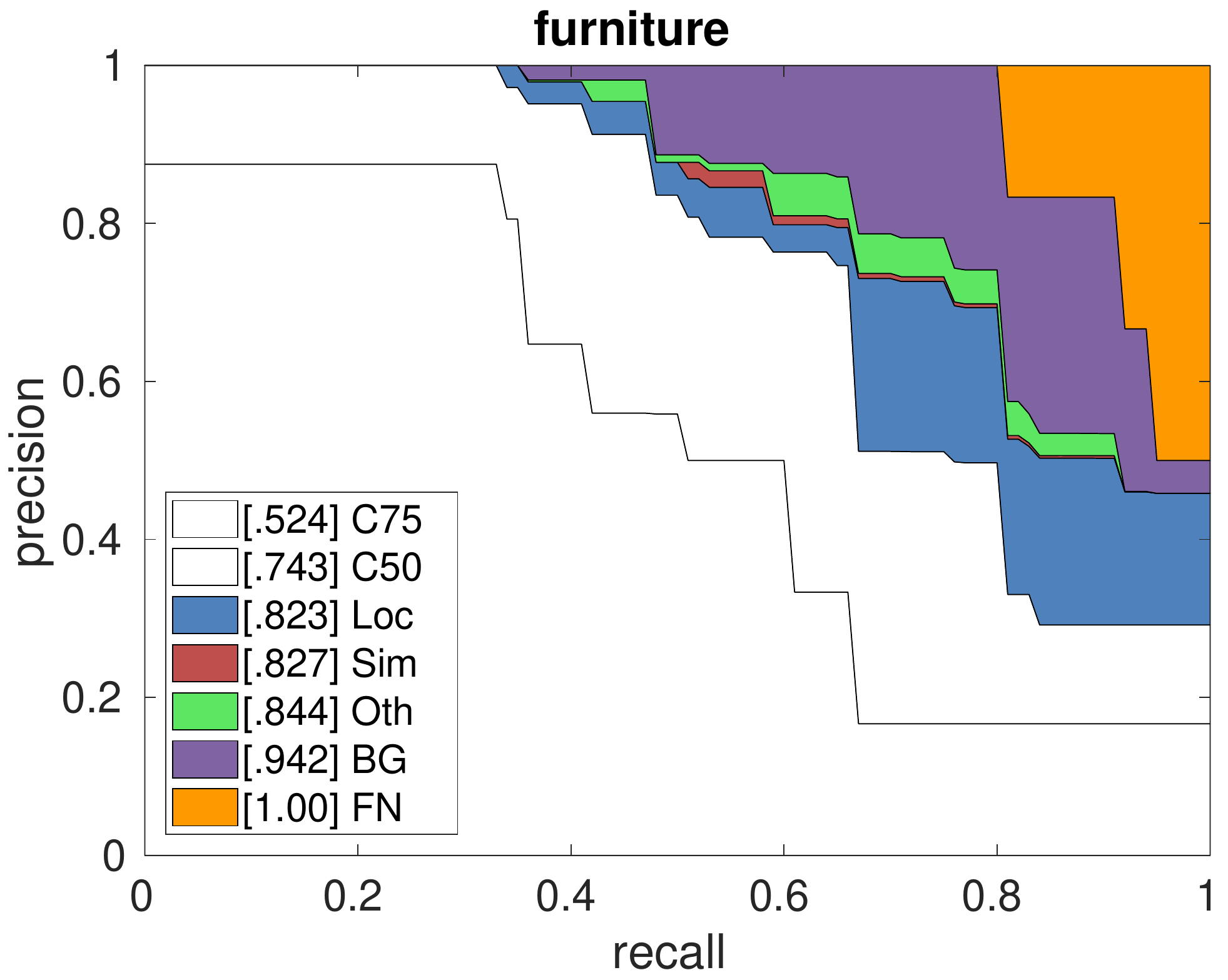}
}
\subfigure { \label{fig:error_electronic}
\includegraphics[width=4cm]{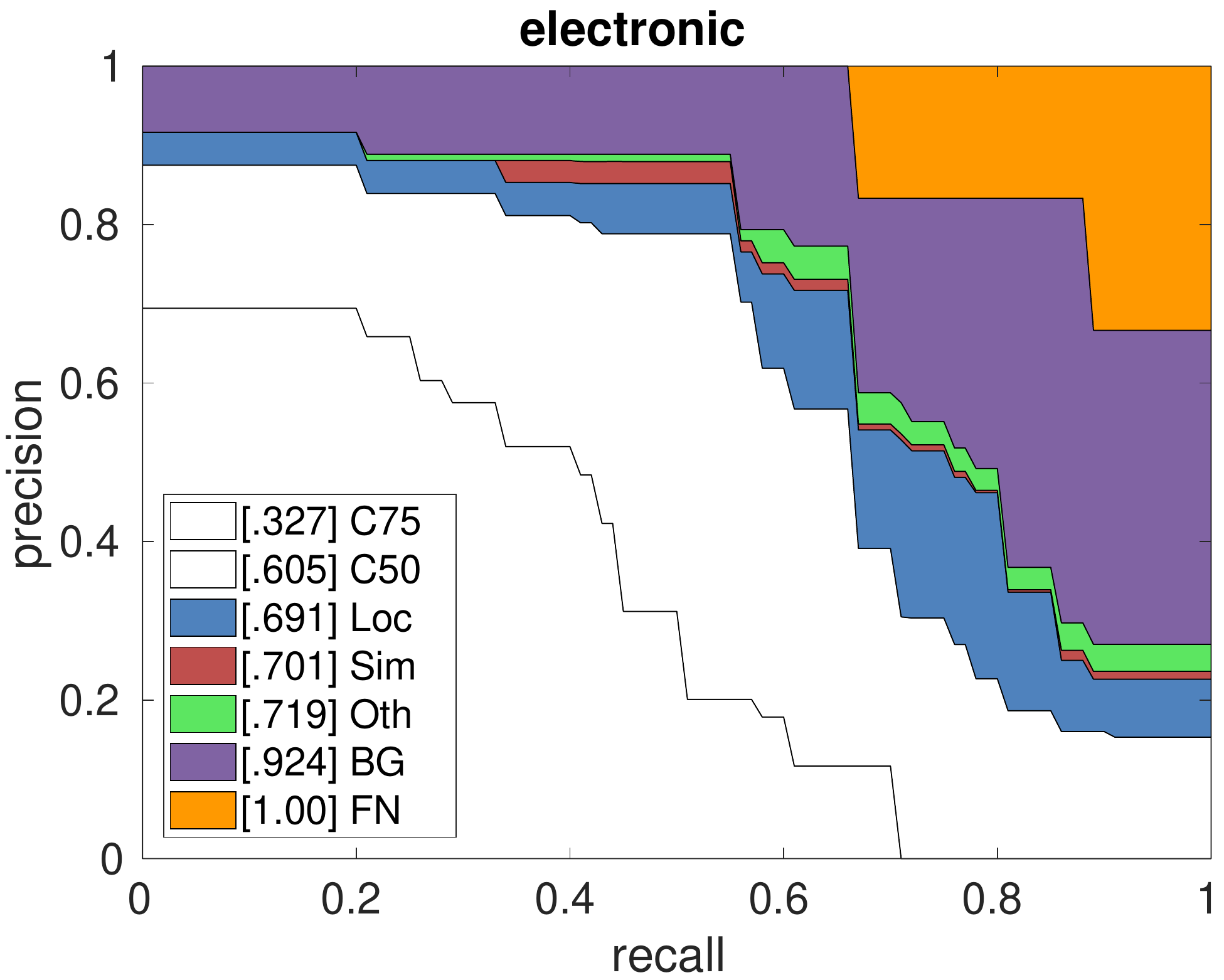}
}
\caption{Error analysis of DRNet512 on person, vehicle, furniture, and electronic classes in the COCO 2014 minival set. Each sub-figure shows the cumulative fraction of detections that are correct (Cor) or false positive due to poor localization (Loc), confusion with similar categories (Sim), with others (Oth), or with background (BG).}
\label{fig:error}
\end{figure}

\subsection{Results on COCO}
We perform a thorough analysis on COCO detection dataset, which contains $80$ class labels. As in previous work, we also use the union of training images and a subset of validation images ($118,278$ images, denoted as ``trainval35k'') for training, and test models on test-dev set ($20,288$ images). The whole network is trained for $70$ epochs with a learning rate of $0.001$, then for $30$ epochs with a learning rate of $0.0001$ and another $30$ epochs with a learning rate of $0.00001$. The main COCO metric denotes as AP, which evaluates detection results at IoU$\in[0.5:0.05:0.95]$. AP@0.5, AP@0.75, AP$_S$, AP$_M$, and AP$_L$ are also used for deep comparison.

As shown in Table~\ref{tab:COCO}, some anchor-free methods recently achieve high AP on COCO, which leverage key-point technology \cite{bib:CornerNet,bib:ExtremeNet,bib:CenterNet} or full convolution \cite{bib:FCOS} for object detection. DRNet320 achieves the results of $30.5\%$, which is better than contemporary methods (e.g., RefineDet320, RFBNet300), so our approach can effectively cope with a variety of complex situations with small input resolution. Furthermore, DRNet512 obtains a more competitive AP of $34.3\%$. Because they have similar AP results, we draw readers' attention to a deep comparison among methods in boldface. (Note that YOLOv3 is not written in bold because its input size is $608$.) At first, DRNet512 has huge improvements as opposed to RefineDet512 on all criteria, where our designs are proved to be successful. Moreover, our DRNet512 has the best VOC-like AP@0.5 (i.e., $57.1\%$) and AP$_S$ (i.e., $17.6\%$), so our method is more adept at small object detection owing to the proposed dual refinement mechanism. However, our results on AP@0.75 and AP$_L$ are not comparable with that of some methods. This is caused by two reasons: i) two-stage methods use larger input size; ii) ResNet101 or RFB block \cite{bib:RFB} provides larger effective receptive field for describing large objects \cite{bib:ERF}. Therefore, our methods are also tested with ResNet101 as the backbone. As a result, more competitive results are induced, i.e., DRNet320-ResNet101 delivers $33.5\%$ AP and DRNet512-ResNet101 achieves $38.6\%$ AP. If multi-scale testing is employed, we see $42.4\%$ AP from DRNet512-ResNet101. Additionally, using MobileNet \cite{bib:MobileNet} as the backbone, our DRNet outperforms Faster RCNN, SSD, and RFBNet by a substantial margin. DRNet and RFBNet are similar in VOC mAP and COCO AP. Although similar performances are produced, DRNet and RFBNet are designed based on different motivations. That is, DRNet solves the problem of inaccurate anchors and feature locations, while RFBNet enhances the receptive field of the backbone. Two ideas are complementary so that anchor-offset detection and RFB block can be employed simultaneously.

Error analysis of DRNet512 is conducted on COCO 2014 minival set ($5,000$ images), and precision-recall curves are shown on person, vehicle, furniture, and electronic classes. From Fig.~\ref{fig:error}, it is seen that there exists room for improvement of location precision. As for classification, DRNet has less confusion with similar categories or others (Sim \& Oth). Thus, our approach is good at inter-class inference, benefiting from accurate single-stage detection features generated by the feature location refinement. By contrast, the error caused by the background (BG) is slightly serious. Probable improvement proposals will be discussed in Section~\ref{sec:Dis}.

\begin{figure}[!t]
 \centering
\includegraphics[width=8cm]{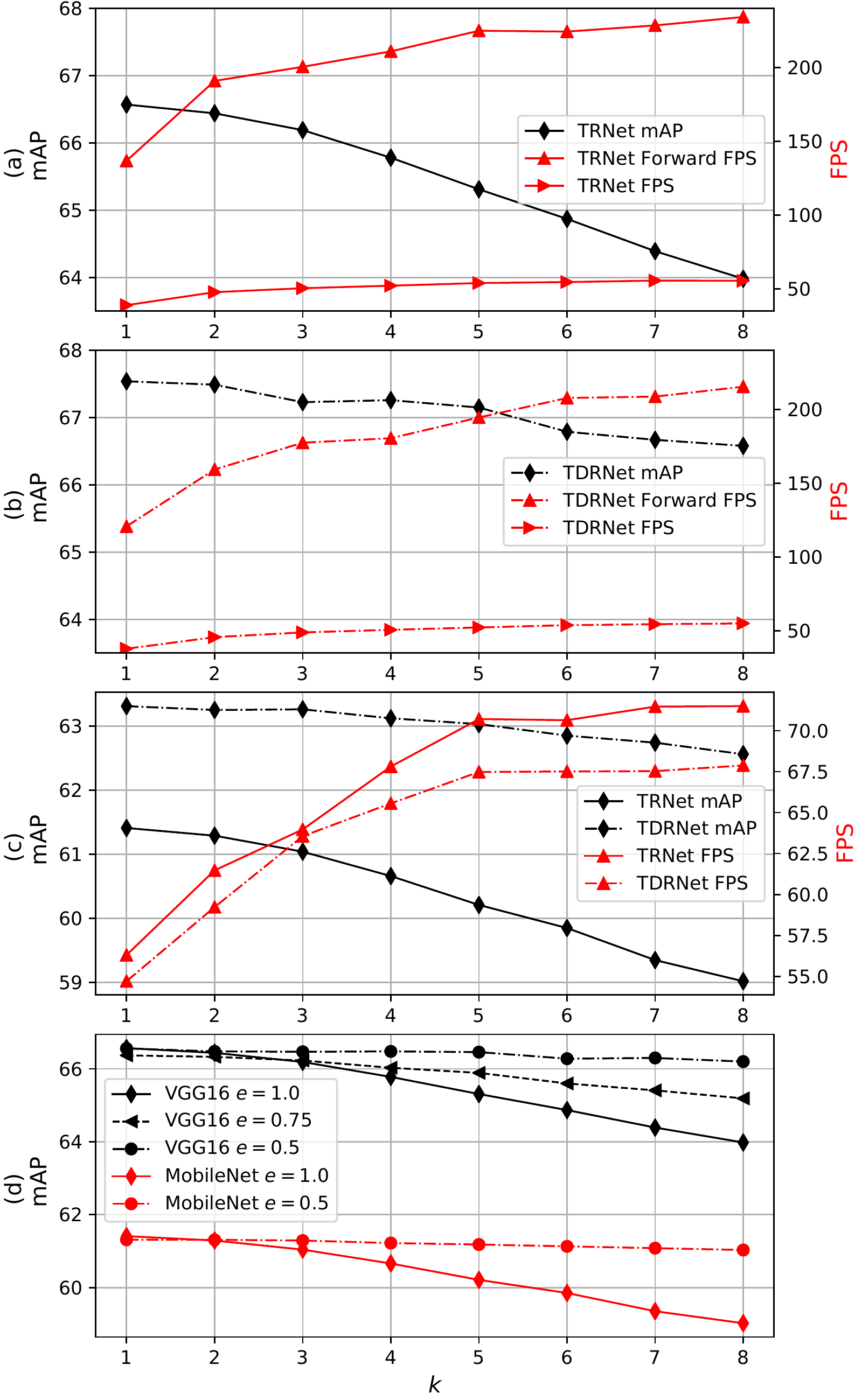}   
\caption{Inference analysis of TRNet and TDRNet with detection period $k$ and soft coefficient $e$. \emph{Forward FPS} does not consider the time consumption caused by NMS. (a) TRNet-VGG16 (The baseline is $63.0\%$). (b) TDRNet-VGG16 (The baseline is $63.0\%$). (c) TRNet and TDRNet with MobileNet (The baseline is $58.3\%$). (d) TRNet with VGG16 and MobileNet.}
\label{fig:kvsmAP}
\end{figure}

\subsection{Results on ImageNet VID}
TRNet and TDRNet are evaluated on ImageNet VID dataset, which requires algorithms to detect $30$-class targets in consecutive frames. There are $4,000$ videos in the training set ($1,181,113$ frames), and $555$ videos in the validation set ($176,126$ frames). The initial learning rate is $0.001$ for the first $70$ epochs, then we use a learning rate of $0.0001$ for the next $30$ epochs and $0.00001$ for another $30$ epochs. For fast inference speed, all models use $320\times 320$ input images.

\subsubsection{Accuracy vs. Speed Trade-Off}
SSD with 4-scale detection features serves as the baseline, called SSD4s, and RG and RD are also contrasted with similar structure (see Fig~\ref{fig:TRN_arch}). As a result, SSD4s-VGG16 and SSD4s-MobileNet obtain $63.0\%, 58.3\%$ in mAP, respectively. The key frame duration is used for temporal detection, so accuracy vs. speed trade-off based on $k$ is first analyzed. As shown in Fig.~\ref{fig:kvsmAP}(a), TRNet significantly improves the mAP by $3.6\%$ (i.e., $66.6\%$ vs. $63.0$). As $k$ increasing, the mAP decreases while the speed raises. Note that NMS impacts detection speed to some extent, but this part is out of the scope of this paper, so the FPS without NMS is also reported (denoted as \emph{Forward FPS}). As plotted in Fig.~\ref{fig:kvsmAP}(a), \emph{Forward FPS} increases from $136.8$ to $234.2$ with the rise of $k$, and the overall speed can reach $55.5$ FPS. Furthermore, TDRNet improves the performance up to $67.5\%$ benefiting from the proposed anchor-offset detection, which outperforms the baseline by $4.5$ points. When $k=8$, TDRNet can run at $55.1$ FPS (\emph{Forward FPS} reaches $215.4$) while maintaining the mAP of $66.6\%$. As for $k=1,2,...,8$, TRNet has a $2.6$-point drop in mAP (i.e., $66.6\%$ vs. $64.0\%$), whereas TDRNet only has a decrease of $0.9\%$ mAP (i.e., $67.5\%$ vs. $66.6\%$). Thus, the refinement information in TDRNet is more robust in terms of temporal propagation owing to our proposed anchor-offset detection. Additionally, using MobileNet as the backbone, TRNet and TDRNet achieve $60.7\%$ and $63.1\%$ mAP ($k=4$), surpassing the baseline by $2.4$ and $4.8$ points, respectively. Meanwhile, our MobileNet-based model can run over $70$ FPS (Noted that the speed of MobileNet in Pytorch is slightly slower than the official implementation).

\begin{figure}[!t]
 \centering
\includegraphics[width=6.5cm]{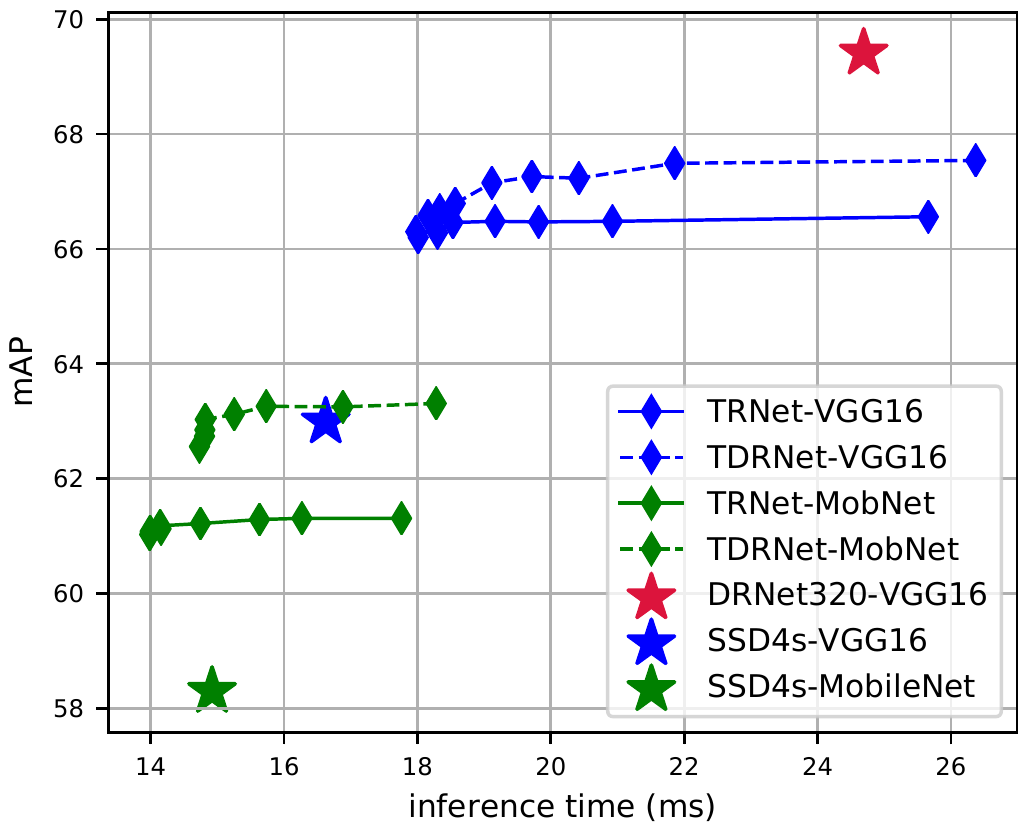}   
\caption{The plot of mAP vs. inference time for approaches in this paper. We achieve a wide variety of trade-offs between accuracy and speed through different backbones and $k$ settings.}
\label{fig:msvsmAP}
\end{figure}

To overcome TRNet's rapid mAP decrease with increasing $k$, the soft refinement strategy is introduced to TRNet with a soft coefficient $e$. Referring to Fig.~\ref{fig:kvsmAP}(d), $e=0.5$ can restrict this mAP drop within $1\%$ from $k=1$ to $k=8$, i.e., $66.6\%$ vs. $66.2\%$ for VGG16 and $61.3\%$ vs. $61.0\%$ for MobileNet.

As shown in Fig.~\ref{fig:msvsmAP}, with $320\times 320$ input size, this paper presents a series of approaches for the trade-off between accuracy and speed. The fast solution is TRNet-MobileNet ($k=8$), whose inference time is $14$ ms. The most accurate method in this paper is DRNet320 with $69.4\%$ mAP and $25$ ms in inference. In terms of TDRNet-VGG16 and DRNet320-VGG16, it can be seen that DRNet is more accurate owing to FPN and multi-deformable head, yet TDRNet has a better trade-off between accuracy and speed.

\begin{table}[!t]
\renewcommand{\arraystretch}{1.1}
\setlength\tabcolsep{1.8pt}
\caption{{\bf Comparison of the proposed methods and several prior and contemporary approaches on VID}. $k=4$ for TRNet and TDRNet.}
\label{tab:VID}
\centering
\begin{tabular}{c| c c c c | c c  }
\Xhline{1.5pt}
\multirow{2}{*}{Method} &          \multicolumn{4}{c|}{Components} &  \multicolumn{2}{c}{Performances}\\
                                 &Backbone   & Flow      & Tracking  & RNN    & Real time     & mAP\\
\hline
\emph{static methods}            &           &           &           &          &               &\\
SSD4s320                         &MobileNet  &           &           &          &\checkmark     & 58.3\\
SSD4s320                         &VGG16     &           &           &          &\checkmark     & 63.0\\
Faster RCNN \cite{bib:FasterRCNN}&GoogLeNet  &           &           &          &               & 63.0\\
SSD300 \cite{bib:SSD}            &VGG16     &           &           &          &\checkmark     & 63.0\\
RefineDet320 \cite{bib:RefineDet}&VGG16     &           &           &          &\checkmark     & 66.7\\
\hline
DRNet320                           &VGG16     &           &           &          &\checkmark     & 69.4\\
\hline
\hline
\emph{offline methods} &&&&&& \\
STMN\cite{bib:STMN}              &VGG16     &           &           &\checkmark&               & 55.6 \\
TPN \cite{bib:TPN}               &GoogLeNet  &           &           &\checkmark&               & 68.4 \\
FGFA\cite{bib:FGFA}              &ResNet101 &\checkmark &           &          &               & 76.3 \\
HPVD \cite{bib:HPVD}             &ResNet101 &\checkmark &           &          &               & 78.6 \\
STSN \cite{bib:STSN}             &ResNet101 &           &           &          &               & 78.7 \\
\hline
\hline
\emph{online methods} &&&&&&\\
LSTM-SSD \cite{bib:LSTM_SSD}      &MobileNet  &          &          &\checkmark &\checkmark     & 54.4 \\
HPVD-Mob \cite{bib:HPVD_Mobiles}&MobileNet&\checkmark&   &\checkmark&\checkmark     & 60.2 \\
TCNN \cite{bib:TCNN}             &DeepID+Craft&\checkmark&\checkmark&           &               & 61.5 \\
TSSD \cite{bib:TSSD}             &VGG16     &           &          &\checkmark&\checkmark      & 65.4 \\
D\&T \cite{bib:DT}               &ResNet101 &           &\checkmark&           &               & 78.7 \\
\hline
TRNet                       &MobileNet  &           &           &          &\checkmark     & 61.2 \\
TDRNet                      &MobileNet  &           &           &          &\checkmark     & 63.1 \\
TRNet                       &VGG16     &           &           &          &\checkmark     & 66.5 \\
TDRNet                      &VGG16     &           &           &          &\checkmark     & 67.3 \\
\Xhline{1.5pt}
\end{tabular}
\end{table}

\subsubsection{Comparison with Other Architectures}
TRNet and TDRNet are compared against several prior and contemporary approaches in Table~\ref{tab:VID}. Existing video detectors are categorized into offline methods (i.e., batch-frame mode) and online methods. Most methods are based on a two-stage detector and a deep backbone, so they usually have high mAP yet impractical execution time. As for offline approaches, this non-causal batch-frame mode usually leverages both previous and future information that prohibits it from real-world applications. In addition, recent works usually borrow other temporal modules (e.g., tracker, optical flow, and LSTM) to integrate multi-frame information. Among single-stage methods, TDRNet-VGG16 has a significant superiority in accuracy, i.e., $1.9\%$ and $12.9\%$ higher mAP than TSSD \cite{bib:TSSD} and LSTM-SSD \cite{bib:LSTM_SSD}, respectively. When compared to MobileNet-based detectors, TDRNet-MobileNet has the best results, i.e., it outperforms LSTM-SSD by $8.7$ points and surpasses HPVD-Mob \cite{bib:HPVD_Mobiles} by $2.9$ points. To the best of our knowledge, our designs have the following merits: i) instead of borrowing other temporal modules, temporal information is exploited from the detector itself. Thus, our design is a new online detection mode for videos; ii) TDRNet achieves the highest mAP among real-time online temporal detectors, and it induces a better trade-off between accuracy and speed for real-world tasks.

\subsection{Underwater Object Detection and Grasping}
Underwater missions are quite intractable for humans, so we use an autonomous system for these difficult tasks, i.e., underwater navigation and object grasping. Owing to the characteristic of good accuracy vs. speed trade-off, TDRNet is suited to real-world applications. In reality, we employ a remote operated vehicle (ROV) for underwater grasping, where a microcomputer with an Intel I5-6400 CPU, an NVIDIA GTX 1060 GPU, and 8 GB RAM is deployed and a camera is placed in the electric compartment for visual navigation. Based on our proposed method, the ROV is able to approach targets and grasp marine products (e.g., sea cucumbers, sea urchins, bivalves, and starfish) using a manipulator. The test venue is located in Zhangzidao, China, where the water depth is approximately $10$ m. It should be noted that the dataset is from \url{http://en.cnurpc.org}. Based on this dataset, we compare some prior approaches and DRNet. Referring to Table~\ref{tab:UWDet}, the proposed DRNet has advantages in accuracy for this difficult real-world mission. Note that Table~\ref{tab:UWDet} does not include TRNet and TDRNet because this is a static dataset.

\begin{table}[!t]
\renewcommand{\arraystretch}{1.1}
\setlength\tabcolsep{4pt}
\caption{{\bf Comparison of the proposed methods and contemporary approaches for underwater object detection.} All these methods use VGG16 backbone and 512 input size}
\label{tab:UWDet}
\centering
\begin{tabular}{c c c c c c}
\Xhline{1.5pt}
Method & mAP & Sea cucumber & Sea urchin & Bivalve & Starfish \\
\hline
SSD \cite{bib:SSD}                & 72.9   & 70.2  & 87.1 & 50.8 & 83.5 \\
RetinaNet \cite{bib:RetinaNet}    & 74.0   & 69.8  & 88.1 & 54.7 & 83.4 \\
RefineDet \cite{bib:RefineDet}    & 76.0   & 73.8  & 90.2 & 54.1 & 85.8 \\
DRNet                               &\bf 77.1   &\bf 75.6  &\bf 91.1 &\bf 55.1 &\bf 86.7 \\
\Xhline{1.5pt}
\end{tabular}
\end{table}

\begin{figure} \centering
\subfigure { \label{fig:rov}
\includegraphics[width=8cm]{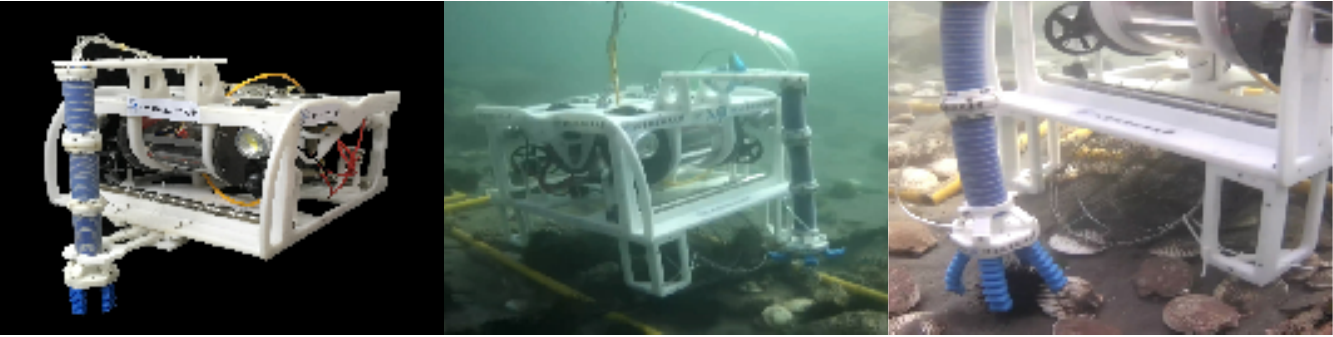}
}
\subfigure { \label{fig:uw}
\includegraphics[width=8cm]{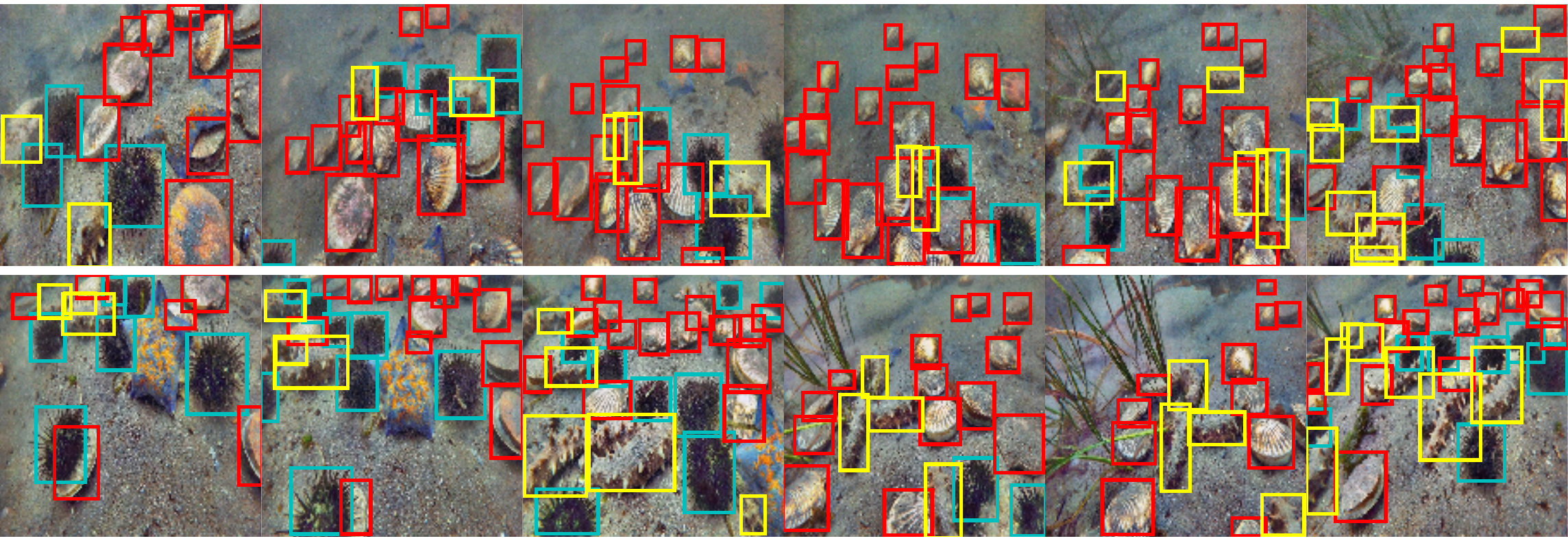}
}
\caption{Schematic examples of an underwater detection/grasping task. TDRNet is trained for detecting seafood animals, i.e., sea cucumbers, sea urchins, and bivalves, which are shown in yellow, cyan, red boxes, respectively. We draw all detected boxes with $>0.4$ score. The top line: The employed ROV and working scenarios; The bottom line: Detection snapshots.}
\label{fig:uw_rov}
\end{figure}

As shown in Fig.~\ref{fig:rov}, the ROV works on a natural seabed, and TDRNet is competent in detecting objects in an unstructured undersea environment (see Fig.~\ref{fig:uw}). For better visualization, we demonstrate the detection results of sea cucumbers, sea urchins, and bivalves using yellow, cyan, red boxes, respectively. This task is challenging for object detection. On one hand, objects gather together and occlude each other. For example, in the left-bottom of $7$th demonstrated frame, a bivalve is almost completely occluded by a sea urchin. On the other hand, many small objects appear in the practical scenario. Despite these difficulties, our proposed TDRNet can deal with them efficiently and demonstrate a promising robotic application. The experimental video is available at \url{https://youtu.be/XDSa4BQX9M8}.

\subsection{Discussion}
\label{sec:Dis}
\subsubsection{Key Frame Scheduling}
Zhu \emph{et al.} developed an adaptive key frame scheduling \cite{bib:HPVD} for key frame selection, and we employ a pre-fixed key frame duration. We argue that the adaptive key frame scheduling is needless for both accuracy and speed in this paper: i) any scheduling strategy cannot generate an mAP that outmatches the result of $k=1$. Therefore, given that the speed of RG is fast enough (i.e., $270$ FPS) and a scheduling strategy should deal with each frame, it is better to set $k=1$ than to use an adaptive key frame scheduling. ii) the longest period is $k=8$ in our experiments, and we also state that longer detection period is needless: \emph{Forward FPS} of SSD4s-VGG16 is $270$ and that of TRNet-VGG16 reaches $234$ ($k=8$). Thus, longer key frame duration has an ignorable contribution to inference speed since TRNet and TDRNet cannot overpass SSD4s in \emph{Forward FPS}.

\subsubsection{Further Enhancement of Refinement Networks}
In terms of accuracy, it can be seen from Fig.~\ref{fig:error} that there still exists room for improvement of location precision and foreground-background classification. We present two probable solutions: i) multi-step refinement could be beneficial; ii) because of the hard negative mining, only a part of negative samples (i.e., background) are used for training. Therefore, using a focal loss \cite{bib:RetinaNet} to train a network with all negative samples could be more effective. For example, Chi \emph{et al.} used focal loss and negative anchor filtering to train a refinement network and achieved high performance on face detection \cite{bib:SelectRefine}.

Regarding inference speed, the NMS has an impact. There could exist two solutions: i) decreasing the anchor amount could be beneficial; ii) an end-to-end detector is becoming urgently necessary. For example, Hu \emph{et al.} developed a relation network for both detection and duplicate removal, so the whole network can perform in an end-to-end manner~\cite{bib:Relation}.

\subsubsection{Challenging Object Detection in Real-World Scenes}
Real-world object detection is more challenging, but classical datasets, e.g., VOC, COCO, and VID, could not contain various real-world environments. Thus, it important to study object detection in specific real-world scenes. To this end, many real-world datasets are developed. For example, besides our studied underwater dataset, TU-VDN dataset \cite{bib:TU-VDN} introduces adverse weather conditions to object detection.

\section{Conclusion and Future Work}
\label{sec:CON}
In this paper, we have taken aim at precisely detecting objects in real time for static and temporal scenes. Firstly, drawbacks of the single-stage detector are analyzed from the strengths of two-stage methods. Thereby, including an anchor refinement, a feature location refinement, and a deformable detection head, a novel anchor-offset detection is proposed. Besides two-step regression, the anchor-offset detection is also able to capture accurate single-stage features for detection. Correspondingly, a DRNet is proposed based on the anchor-offset detection, where a multi-deformable head is also designed for more contextual information. In the case of temporal detection, we propagate the refinement information in the anchor-offset detection across time and propose a TRNet and a TDRNet with a reference generator and a refinement detector. Our developed approaches have been evaluated on PASCAL VOC, COCO, and ImageNet VID. As a result, our designs induce a considerably enhanced detection accuracy and see a substantial improvement on the trade-off between accuracy and speed. Finally, the proposed algorithms are applied to underwater object detection and grasping.

In the future, we plan to incorporate attention mechanism to the anchor-offset detection and design more effective networks for more robust feature learning.

\begin{table*}[!t]
\renewcommand{\arraystretch}{1.0}
\setlength\tabcolsep{2pt}
\caption{AP list on PASCAL VOC test set by the proposed methods. All models use VGG16 as the backbone network.}
\label{tab:all_VOC_AP}
\centering
\begin{tabular}{c c c c c c c c c c c c c c c c c c c c c c}
\Xhline{1.5pt}
Method & aero &bike  &bird  &boat  &bottle& bus  &car   &cat   &chair &cow   &table  &dog  &horse &mbike &person&plant &sheep &sofa  &train &tv &mAP\\
\hline
\emph{VOC 2007}&  &          &       &         &       &       &       &        &       &       &   \\
DRN320 &86.31 &87.45 &83.01 &77.35 &66.12 &87.95 &88.64 &89.44 &69.30 &85.73 &76.34 &87.50 &88.99 &87.16 &83.94 &57.48 &85.24 &82.28 &88.07 &82.12 &82.02\\
DRN512 &88.83 &86.52 &85.45 &77.50 &72.23 &88.05 &89.03 &89.92 &68.58 &88.11 &76.81 &87.50 &89.30 &85.50 &85.44 &59.18 &85.95 &80.75 &87.30 &82.99 &82.75\\
DRN320+&89.82 &88.29 &86.37 &81.26 &73.26 &88.52 &89.48 &89.21 &71.69 &88.50 &76.88 &88.12 &89.48 &89.10 &86.01 &62.59 &86.39 &83.02 &88.09 &82.81 &83.94\\
DRN512+&90.38 &88.42 &86.08 &79.79 &76.82 &89.03 &89.51 &90.15 &72.87 &88.31 &81.57 &88.34 &89.80 &87.63 &86.56 &62.16 &86.71 &82.12 &88.24 &83.52 &84.40\\
\hline
\emph{VOC 2012}&  &          &       &         &       &       &       &        &       &       &   \\
DRN320 &88.92 &87.12 &80.36 &69.18 &60.23 &84.49 &82.77 &92.42 &62.27 &83.33 &66.76 &91.08 &88.01 &87.21 &86.79 &57.79 &82.97 &72.09 &88.04 &75.07 &79.34\\
DRN512 &91.57 &88.00 &83.27 &69.42 &68.45 &85.53 &85.53 &93.16 &62.10 &85.51 &65.40 &91.69 &87.89 &88.51 &88.56 &59.68 &86.68 &66.64 &88.98 &75.83 &80.60\\
DRN320+&91.16 &89.38 &85.56 &75.18 &70.50 &87.49 &88.01 &93.60 &66.77 &85.84 &69.42 &92.87 &89.54 &90.13 &90.14 &66.16 &86.65 &73.51 &90.17 &79.75 &83.09\\
DRN512+&92.20 &89.82 &86.73 &74.16 &72.86 &86.91 &88.76 &94.21 &66.30 &88.21 &68.40 &93.54 &90.11 &89.80 &90.72 &64.69 &89.70 &72.88 &90.86 &80.45 &83.58\\
\Xhline{1.5pt}
\end{tabular}
\end{table*}

\begin{table*}[!t]
\renewcommand{\arraystretch}{1.0}
\setlength\tabcolsep{8.5pt}
\caption{Object detection results on the MS COCO 2015 test-dev set.}
\label{tab:all_COCO_AP}
\centering
\begin{tabular}{c c c c c c c c c c c c c c c c c c c c c c}
\Xhline{1.5pt}
Method    & AP  &AP@0.5 &AP@0.75 &AP$_S$ &AP$_{M}$ &AP$_{L}$ &AR$_{1}$ &AR$_{10}$ &AR$_{100}$ &AR$_{S}$ & AR$_{M}$ & AR$_{L}$ \\
\hline
DRN320-MobileNet&26.0 &45.3   &26.8    &8.0   &28.7  &38.9     &24.1     &36.4      &38.4       &13.0     &44.2      &56.9 \\
DRN512-MobileNet&28.5 &49.8   &29.6    &14.3  &32.1  &36.6     &25.5     &39.9      &42.2       &22.2     &47.3      &55.0 \\
DRN320-VGG16&30.5 &51.2   &32.3    &11.2   &33.9     &44.9     &27.0     &40.9      &42.9       &17.5     &49.0      &62.2 \\
DRN512-VGG16&34.3 &57.1   &36.4    &17.9   &38.1     &44.8     &28.9     &44.8      &47.4       &26.6     &52.7      &61.5 \\
DRN320+-VGG16&35.4 &57.8   &37.7    &19.7   &38.5     &45.7     &30.2     &47.1      &49.7       &30.2     &54.5      &63.2 \\
DRN512+-VGG16&37.9 &61.6   &40.3    &22.6   &40.4     &48.4     &31.4     &49.4      &52.3       &34.6     &55.5      &65.5 \\
DRN320-ResNet101 &33.5 &53.4   &35.9    &11.5   &37.4     &50.6     &29.2     &43.6      &45.7       &18.1     &51.7      &67.5 \\
DRN512-ResNet101 &38.6 &60.3   &42.2    &19.0   &43.2     &52.7     &31.9     &48.8      &51.3       &27.7     &57.3      &69.3 \\
DRN320+-ResNet101 &39.2 &61.3   &42.3    &21.4   &42.8     &51.4    &32.7     &50.5      &53.2       &31.8     &58.5      &69.0 \\
DRN512+-ResNet101 &42.4 &65.5   &46.1    &25.7   &45.3     &55.0    &34.2     &53.5      &56.5       &37.8     &60.1      &71.9 \\
\Xhline{1.5pt}
\end{tabular}
\end{table*}

\begin{table*}[!t]
\renewcommand{\arraystretch}{1.0}
\setlength\tabcolsep{3pt}
\caption{AP list on ImageNet VID 2017 validation set by the proposed methods. All models use $320\times 320$ input images and are trained with VID+DET dataset. $k=4$ in TRN and TDRN inference.}
\label{tab:all_VID_AP}
\centering
\begin{tabular}{c c c c c c c c c c c c c c c c c}
\Xhline{1.5pt}
Method    &airplane &antelope &bear  &bicycle &bird  & bus  & car  &cattle & dog  &d.cat &elephant& fox  &g.panda &hamster &horse & lion \\
\hline
DRN320    &87.56    &80.96    &71.18 &69.63   &69.25 &68.53 &64.12 &73.03  &55.93 &65.48 &73.73   &85.63 &81.84   &89.78   &69.86 &34.89  \\
SSD4s-VGG16&83.45   &71.62    &66.16 &57.23   &63.02 &73.13 &58.79 &58.05  &54.88 &66.39 &70.56   &79.30 &80.67   &86.00   &64.50 &38.52\\
TRN-VGG16 &83.38    &69.42    &71.54 &60.44   &63.93 &69.74 &58.67 &58.64  &54.84 &71.60 &69.96   &84.79 &81.15   &89.28   &69.33 &39.72\\
TDRN-VGG16&83.67    &73.80    &71.48 &60.89   &64.09 &75.22 &59.94 &60.81  &57.49 &70.59 &71.39   &84.55 &81.93   &88.47   &68.42 &46.49\\
SSD4s-MobileNet&81.35&71.27   &67.15 &52.10   &59.28 &66.15 &58.23 &54.87  &46.35 &61.61 &65.50   &73.79 &79.50   &75.12   &59.36 &9.19\\
TRN-MobileNet &80.87&72.99    &67.23 &56.10   &62.54 &68.61 &59.46 &61.16  &48.77 &62.51 &67.34   &72.43 &78.94   &83.54   &65.54 &15.54\\
TDRN-MobileNet&82.67&76.91    &62.92 &58.67   &63.97 &69.10 &59.91 &61.50  &53.17 &67.55 &65.98   &74.74 &80.77   &85.29   &65.55 &11.89\\
\hline
Method    &lizard &monkey &m.bike &rabbit &r.panda &sheep &snake &squirrel &tiger &train &turtle &w.craft &whale &zebra &\multicolumn{2}{c}{mAP} \\
\hline
DRN320    &71.40  &43.20  &80.79  &52.99  &62.35   &60.56 &52.63 &47.76    &89.16 &82.82 &76.24  &62.03   &69.81 &89.24  & \multicolumn{2}{c}{69.41}\\
SSD4s-VGG16&60.56 &40.76  &75.68  &45.47  &16.77   &48.88 &44.48 &46.14    &84.27 &76.26 &70.46  &66.23   &57.43 &83.95  & \multicolumn{2}{c}{62.99}\\
TRN-VGG16 &69.31  &42.58  &76.78  &53.03  &55.05   &51.35 &53.36 &47.61    &86.34 &78.55 &73.31  &62.25   &63.11 &85.50  & \multicolumn{2}{c}{66.49}\\
TDRN-VGG16&72.63  &42.36  &77.43  &56.59  &41.32   &58.49 &45.75 &50.82    &85.16 &81.13 &74.36  &62.30   &65.53 &84.82  & \multicolumn{2}{c}{67.26}\\
SSD4s-MobileNet&54.17&33.51&72.19 &41.43  &21.81   &51.51 &35.60 &41.47    &83.72 &75.54 &63.05  &53.73   &57.53 &82.81  & \multicolumn{2}{c}{58.30}\\
TRN-MobileNet &55.25&36.65&73.87  &55.19  &29.41   &52.22 &40.69 &44.88    &83.61 &73.71 &69.78  &53.54   &59.93 &84.55  & \multicolumn{2}{c}{61.23}\\
TDRN-MobileNet&58.36&37.54&77.13  &58.39  &49.43   &50.24 &35.56 &45.05    &84.07 &77.45 &71.97  &59.74   &63.85 &84.42  & \multicolumn{2}{c}{63.13}\\
\Xhline{1.5pt}
\end{tabular}
\end{table*}

\end{document}